\documentclass{article}

\PassOptionsToPackage{numbers,sort&compress}{natbib}
\usepackage[eandd, final]{neurips_2026}

\usepackage[utf8]{inputenc}
\usepackage[T1]{fontenc}
\usepackage{float}
\usepackage{hyperref}
\usepackage{url}
\usepackage{booktabs}
\usepackage{amsfonts}
\usepackage{amsmath}
\usepackage{amssymb}
\usepackage{nicefrac}
\usepackage{microtype}
\usepackage{xcolor}
\usepackage{graphicx}
\usepackage{enumitem}
\usepackage{comment}

\definecolor{ciText}{RGB}{78,132,146}
\newcommand{\ci}[1]{\,{\textcolor{ciText}{\scriptsize #1}}}
\graphicspath{{figures/}}

\title{OpenBrain: An Auditable Generated-Label Release for Whole-Brain MRI Parcellation}

\author{%
  \textbf{Qizhen Lan$^{1}$ \quad Yu-Chun Hsu$^{1}$ \quad Yuxiang Wei$^{2}$ \quad Lijing Zhu$^{3}$} \\
  \textbf{Zenan Sun$^{1}$ \quad Liang He$^{4}$ \quad Lishan Yu$^{1}$ \quad Xiaoqian Jiang$^{1}$}\thanks{Corresponding author: \href{mailto:Xiaoqian.Jiang@uth.tmc.edu}{\texttt{Xiaoqian.Jiang@uth.tmc.edu}}. Contact for the data and code: \href{mailto:Qizhen.Lan@uth.tmc.edu}{\texttt{Qizhen.Lan@uth.tmc.edu}}.} \\[0.5em]
  {\normalfont\small $^{1}$The University of Texas Health Science Center at Houston} \\
  {\normalfont\small $^{2}$Georgia Institute of Technology} \\
  {\normalfont\small $^{3}$University of Houston--Clear Lake} \\
  {\normalfont\small $^{4}$ETH Zurich} \\
}

\begin{document}

\maketitle

\begin{abstract}
Automatic tools can now generate whole-brain MRI parcellations at cohort scale,
but generated-label releases often provide little evidence about which cases
are reliable enough to reuse. We introduce \textbf{OpenBrain}, a
confidence-aware release of 35{,}838 provenance-verified OpenNeuro T1w cases
from 607 source datasets. OpenBrain is organized as an auditable case-record
release rather than a directory of standalone segmentation files: each record
links a primary parcellation, two committee parcellations, source/license
provenance, raw QC-7 measurements, percentile risks, a combined
risk score $R$, and a confidence tier. QC-7 is a reference-free reliability
panel over label geometry, image-label consistency, and committee disagreement,
so users can filter, weight, audit, or re-threshold generated labels without
rerunning inference. We evaluate this released evidence under a frozen
protocol. Across external cohorts, lower QC-7 risk corresponds to higher
generated-label quality; in fixed-budget training, lower-risk selections
improve paired macro Dice over random selection by $+3.2$ to $+3.7$ percentage
points. OpenBrain also
exposes raw axes, committee outputs, provenance, and datasheet documentation, so
confidence claims remain inspectable after download. It does not turn generated
labels into manual annotations or clinical-grade segmentations. Its contribution
is a large open generated-parcellation substrate, a reusable case-level
risk-evidence layer, and an external validation protocol for release-time
QC. \footnote{Availability. Dataset:
\url{https://huggingface.co/datasets/lanqz7766/OpenBrain}. Code:
\url{https://github.com/lanqz7766/openbrain}.}
\end{abstract}

\section{Introduction}
\label{sec:intro}

Whole-brain MRI parcellation has reached a point where automatic labels can be generated much faster than humans can review them. Public T1-weighted MRI repositories such as OpenNeuro~\citep{markiewicz2021openneuro} provide
large, heterogeneous source cohorts, while automatic segmentation models such as FastSurfer~\citep{henschel2020fastsurfer}, QuickNAT~\citep{roy2019quicknat},
and SynthSeg/SynthSeg+~\citep{billot2023synthseg,billot2023synthsegplus}
make cohort-scale anatomical labeling practical. That scale creates a release-time problem. Once generated labels are distributed for reuse, downstream users need more than a plausible segmentation file: they need to trace the source scan, inspect likely failure modes, filter or weight cases, and evaluate reuse without manually reviewing every 3D volume.

Existing resources leave two gaps. First, \emph{scale, diversity, and whole-brain parcellation labels rarely appear in the same open resource}. Manual
benchmarks such as Mindboggle-101~\citep{klein2012mindboggle}, the Hammersmith
atlas~\citep{hammers2003atlas,gousias2008atlas}, LPBA40~\citep{shattuck2008lpba40},
and IBSR~\citep{frazier2007ibsr} remain the appropriate references for
anatomical accuracy, but manual 3D annotation is too time-consuming to scale to tens of thousands of heterogeneous public scans. By contrast, many public segmentation resources provide images, brain masks, tissue labels, lesion labels, or task-specific challenge splits, including pathology-focused benchmarks such as BraTS~\citep{baid2021brats}, ATLAS~\citep{liew2022atlas}, and ISLES~\citep{hernandez2022isles}. 

\begin{figure}[t]
  \centering
  \includegraphics[width=\linewidth]{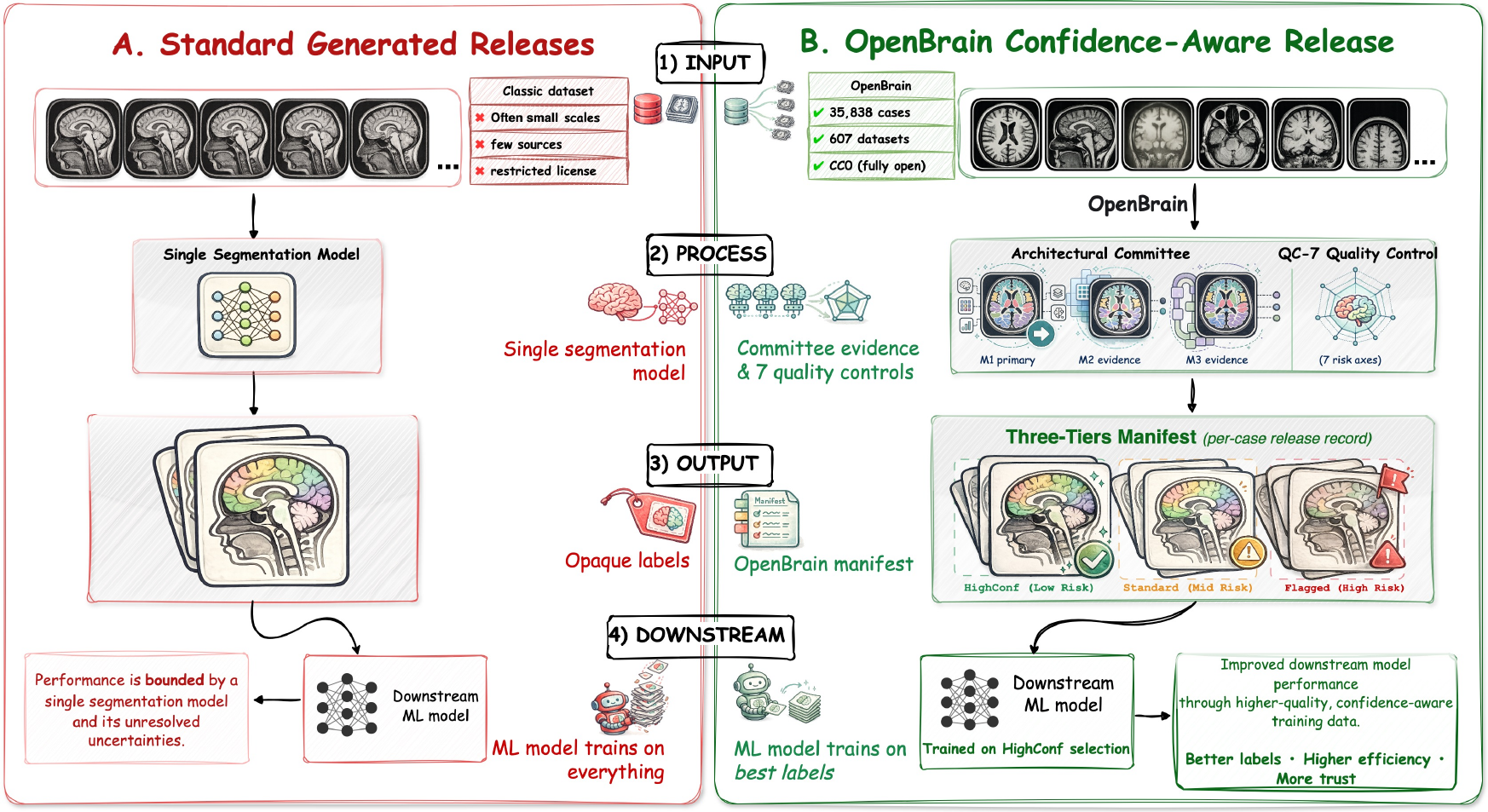}
  \caption{Release-time contrast between a standard generated-label release and
  OpenBrain. OpenBrain packages each parcellation with provenance, committee
  outputs, QC-7 risk fields, confidence tiers, and validation evidence.}
  \label{fig:intro_release_contrast}
\end{figure}

Second, \emph{generated-label releases often lack case-level evidence for reuse.} A released label file by itself does not tell users whether the predicted anatomy has visible failure signs, whether independent label generators agree, or whether the case should be prioritized, down-weighted, audited, or excluded. Reference-free QC and uncertainty methods provide useful diagnostics~\citep{esteban2017mriqc,valindria2017rca,robinson2019rcaukbb,jungo2019uncertainty,wang2019tta},
and large generated-label releases show that automatic labels can scale~\citep{bourigault2025ukbob}. What remains missing is a documentation methodology for
multi-source whole-brain generated labels: what should be released with each
label so that it can be reused and audited after download?

A reusable generated-label release therefore needs three properties: source-scale coverage, case-level reliability evidence, and validation that the released evidence supports the decisions it is meant to inform.
OpenBrain is designed around these three requirements. It releases generated whole-brain parcellations for 35{,}838 provenance-verified T1w cases across 607 OpenNeuro source datasets. The release preserves source identifiers and CC0-compatible public-source provenance, so
derived labels can be traced back to their public sources and reused as a large training and evaluation substrate. Figure~\ref{fig:intro_release_contrast}
summarizes the design shift: OpenBrain releases generated labels as auditable case records rather than as standalone model outputs.

For case-level evidence, OpenBrain releases each case as an
auditable record rather than as a parcellation file alone. The record links the generated whole-brain parcellation to its source provenance and to OpenBrain
QC-7, a compact panel of seven reference-free reliability checks. The QC-7 checks ask three direct questions: does the label geometry look plausible, are the predicted anatomical regions consistent with the image intensities, and do independent parcellation models agree with one another? OpenBrain stores the resulting per-case risk fields and confidence tier in the release case record, so users can filter, weight, audit, or re-threshold cases without rerunning label generation. These fields are not
manual approval labels; they are reusable evidence about where generated labels may be more or less reliable.

We then evaluate whether the released evidence supports the reuse decisions it is meant to inform. Before confirmatory evaluation, the QC definition, percentile-risk conversion, label space, and tier thresholds are fixed. With
these choices fixed, lower-risk retention improves external macro Dice by up to $+3.7$ pp across three external cohorts. In the fixed-budget training test,
lower-risk selections improve paired 96-label macro Dice by $+3.2$ to
$+3.7$ pp over random selection. These results support using OpenBrain's risk fields to rank generated labels and avoid the least reliable tail of the release. OpenBrain makes three contributions:

\begin{enumerate}[leftmargin=1.5em,itemsep=2pt,topsep=2pt]
    \item \textbf{A large multi-source generated-parcellation release.}
    OpenBrain provides 35{,}838 provenance-verified T1w cases from 607
    OpenNeuro source datasets, with generated whole-brain parcellations and
    CC0-compatible provenance metadata (Section~\ref{sec:dataset}).

    \item \textbf{A released risk-evidence layer for generated parcellations.}
    OpenBrain defines and releases QC-7, a reference-free scoring layer that
    converts label-geometry, image-label consistency, and committee-disagreement
    checks into percentile risks and a combined risk score $R$.
    The released risk fields and tiers support filtering, weighting, auditing,
    and re-thresholding without rerunning label generation
    (Section~\ref{sec:dataset:confidence}).

    \item \textbf{External validation of release-time QC evidence.}
    OpenBrain tests whether the released risk fields rank generated-label
    quality on external cohorts and select useful training cases under a fixed
    generated-label budget (Sections~\ref{sec:stage_a}
    and~\ref{sec:training_selection}).
\end{enumerate}

\section{Related Work}
\label{sec:related}

\begin{table}[t]
\centering
\scriptsize
\setlength{\tabcolsep}{2.8pt}
\renewcommand{\arraystretch}{1.08}
\newcommand{\yesmark}{\textcolor{green!60!black}{\ensuremath{\checkmark}}}
\newcommand{\nomark}{\textcolor{red!80!black}{\ensuremath{\times}}}
\newcommand{\partialmark}{\textcolor{orange!85!black}{\ensuremath{\circ}}}
\caption{Positioning of OpenBrain against public brain MRI segmentation
resources. W: whole-brain labels; M: multi-cohort or acquisition diversity; O:
open redistribution; Q: released QC/risk fields or release audit.}
\label{tab:related_brain_segmentation}
\begin{tabular*}{\linewidth}{@{\extracolsep{\fill}} l l l l c c c c l @{}}
\toprule
Benchmark / resource & Scale & MRI / cohort & Labels & W & M & O & Q & Primary role \\
\midrule
\multicolumn{9}{@{}l}{\itshape Brain extraction, tissue, and anatomical segmentation benchmarks} \\
NFBS~\citep{puccio2016nfbs} & 125 scans & NKI T1w & Brain masks & \nomark & \partialmark & \yesmark & \partialmark & Skull stripping \\
CC-359~\citep{souza2018cc359} & 359 scans & Multi-vendor T1w & Brain masks & \nomark & \yesmark & \yesmark & \partialmark & Skull stripping \\
Mindboggle-101~\citep{klein2012mindboggle} & 101 subj. & Multi-source T1w & DKT cortex & \partialmark & \yesmark & \partialmark & \nomark & Cortical anatomy \\
Hammers atlas~\citep{hammers2003atlas,gousias2008atlas} & 30 subj. & Adult T1w & 83 ROIs & \yesmark & \nomark & \partialmark & \nomark & Atlas anatomy \\
LPBA40~\citep{shattuck2008lpba40} & 40 subj. & Adult T1w & 56 ROIs & \yesmark & \nomark & \partialmark & \nomark & Reg./seg. eval. \\
IBSR 1.5mm~\citep{frazier2007ibsr} & 18 subj. & T1w, 7--71y & 43 ROIs & \yesmark & \nomark & \partialmark & \nomark & Legacy gold std. \\
MALC / NM~\citep{landman2012malc} & 30 scans; 25 subj. & OASIS T1w & WB atlas & \yesmark & \nomark & \partialmark & \nomark & Multi-atlas challenge \\
CANDIShare~\citep{kennedy2012candishare} & 103 subj. & Pediatric T1w & CMA labels & \yesmark & \partialmark & \partialmark & \nomark & Pediatric anatomy \\
MRBrainS13~\citep{mendrik2015mrbrains13} & 20 scans; 5/15 split & Elderly 3T MRI & GM/WM/CSF & \partialmark & \partialmark & \partialmark & \nomark & Tissue challenge \\
MRBrainS18~\citep{kuijf2024mrbrains18} & 30 subj. & T1/T1-IR/FLAIR & 8 eval. labels & \partialmark & \partialmark & \yesmark & \nomark & Tissue challenge \\
FeTA~\citep{payette2021feta} & 50 recon. & Fetal T2w & 7 tissues & \yesmark & \partialmark & \partialmark & \nomark & Fetal segmentation \\
iSeg-2017~\citep{wang2019iseg} & 23 infant scans & 6-mo T1/T2 & GM/WM/CSF & \partialmark & \partialmark & \partialmark & \nomark & Infant tissues \\
\midrule
\multicolumn{9}{@{}l}{\itshape Pathology-focused segmentation challenges} \\
BraTS 2021~\citep{baid2021brats} & 2{,}040 pts; 1251/219/570 & Glioma mpMRI & Tumor regions & \nomark & \yesmark & \partialmark & \nomark & Tumor segmentation \\
ATLAS v2.0~\citep{liew2022atlas} & 1{,}271 cases & Stroke T1w & Lesion masks & \nomark & \yesmark & \partialmark & \nomark & Stroke lesions \\
ISLES 2022~\citep{hernandez2022isles} & 400 cases & DWI/ADC/FLAIR & Lesion masks & \nomark & \yesmark & \partialmark & \nomark & Acute stroke \\
WMH Challenge~\citep{kuijf2019wmh} & 60 scans; 3 centers & T1/FLAIR & WMH masks & \nomark & \yesmark & \partialmark & \nomark & WMH lesions \\
MSSEG 2016~\citep{commowick2018msseg} & 53 MS cases; 4 centers & Multi-modal MRI & MS lesions & \nomark & \yesmark & \partialmark & \nomark & MS lesions \\
ISBI 2015 MS~\citep{carass2017isbi} & 82 scans; 21/61 split & Longitudinal MRI & MS lesions & \nomark & \partialmark & \partialmark & \nomark & Long. MS lesions \\
\midrule
\textbf{OpenBrain (ours)} & \textbf{35{,}838 cases} & \textbf{607 OpenNeuro T1w sets} & \textbf{WB auto labels} & \yesmark & \yesmark & \yesmark & \yesmark & \textbf{Scale + provenance + QC} \\
\bottomrule
\end{tabular*}
\end{table}

\paragraph{Manual and human-reviewed segmentation datasets.}
Manual and human-reviewed datasets define anatomical targets and remain the right reference for accuracy measurement. In brain MRI, Mindboggle-101~\citep{klein2012mindboggle},
the Hammersmith atlas family~\citep{hammers2003atlas,gousias2008atlas},
LPBA40~\citep{shattuck2008lpba40}, IBSR~\citep{frazier2007ibsr}, MALC~\citep{landman2012malc}, MRBrainS18~\citep{kuijf2024mrbrains18},
and FeTA~\citep{payette2021feta} provide manual or expert-curated labels at modest scale. Pathology-focused resources such as BraTS~\citep{baid2021brats}, ATLAS~\citep{liew2022atlas}, ISLES~\citep{hernandez2022isles}, WMH Challenge~\citep{kuijf2019wmh}, MSSEG~\citep{commowick2018msseg}, and ISBI MS~\citep{carass2017isbi} provide human-reviewed lesion or tumor annotations for specific tasks. Larger medical segmentation resources such as AbdomenAtlas~\citep{qu2023abdomenatlas8k,li2024abdomenatlas}, TotalSegmentator~\citep{wasserthal2023totalsegmentator}, and
TotalSegmentator-MRI~\citep{hantze2024totalsegmentatormri} expand scale through automatic tools and human review. These resources provide targets, references, or task-specific annotations. OpenBrain addresses a different setting: generated whole-brain parcellations released at scale across many public MRI sources,
where per-case manual approval is unavailable and release-time reliability evidence must accompany every case.

\paragraph{Brain parcellation methods and public MRI resources.}
Brain parcellation tools make generated labels feasible, while public
repositories provide source images.
FreeSurfer~\citep{fischl2002wholebrain,fischl2012freesurfer} established a
widely used anatomical parcellation convention;
FastSurfer~\citep{henschel2020fastsurfer}, QuickNAT~\citep{roy2019quicknat},
and SynthSeg/SynthSeg+~\citep{billot2023synthseg,billot2023synthsegplus} make
whole-brain segmentation faster or more robust across heterogeneous inputs.
Knowledge distillation~\citep{lan2026clockdistill} offers another route to compact segmentation
models~\citep{lan2025acam,lan2026reco,fan2026detail,ding2026displacement}, including models designed for
on-premises clinical use~\citep{lan2026performance}.
OpenNeuro~\citep{markiewicz2021openneuro} provides the public BIDS source
layer, and OpenBHB~\citep{dufumier2022openbhb} provides a large multi-site
brain MRI resource for representation learning and bias analysis. This
ecosystem gives OpenBrain its source layer and label-generation tools, but tool
output plus public images does not by itself define a dataset-level reliability
record. OpenBrain adds provenance, committee parcellations, OpenBrain QC-7 risk
fields, percentile risks, and confidence tiers for 35{,}838
provenance-verified cases from 607 OpenNeuro source datasets.

\paragraph{Reference-free QC and generated-label releases.}
Reference-free QC methods provide diagnostics, but their release role differs from OpenBrain. MRIQC~\citep{esteban2017mriqc} targets raw-image quality rather than generated-label reliability. Reverse-classification accuracy~\citep{valindria2017rca,robinson2019rcaukbb}
estimates segmentation quality without ground truth, but is expensive at large label counts. Test-time augmentation uncertainty~\citep{jungo2019uncertainty,wang2019tta}
exposes model instability, but its interpretation depends on the model and domain. UKBOB~\citep{bourigault2025ukbob} is the closest generated-label release
precedent because it scales generated MRI masks and uses filtering plus subset audit. OpenBrain targets multi-source whole-brain parcellation rather than mask
release, and releases the QC axes, percentile risk scores, confidence tiers, and downstream validation as part of the dataset itself, so risk evidence remains available for reuse after download.

\section{The OpenBrain Release}
\label{sec:dataset}

\subsection{Release scope and case-level record}
\label{sec:dataset:construction}
OpenBrain is a generated-label release attached to existing public T1-weighted
MRI records, not a new collection of human subjects. The submitted strict
hosted release contains 35{,}838
provenance-verified cases from 607 OpenNeuro~\citep{markiewicz2021openneuro}
source datasets; for these cases, the raw T1w inputs exactly match public
OpenNeuro snapshots with CC0/public-domain-compatible license metadata. The
release manifest used in this paper is this hosted 35{,}838-case release. A
release case is one source-scoped T1w scan/session record, not an
identity-linked person across unrelated sources.

\begin{figure}[t]
  \centering
  \includegraphics[width=\linewidth]{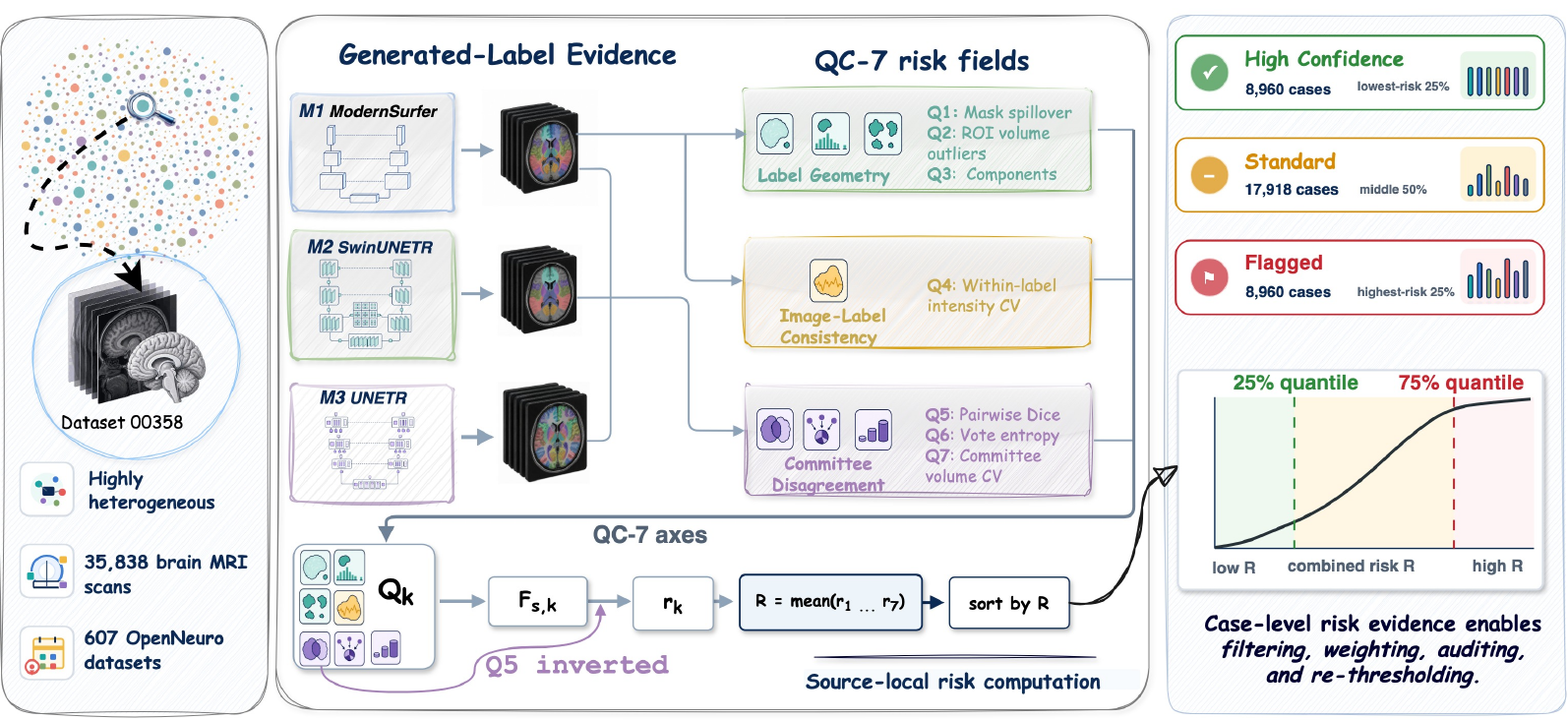}
  \caption{OpenBrain case-level release record. Each source-scoped T1w case
  links generated parcellations, committee outputs, QC-7 checks, percentile
  risks, combined risk $R$, and a confidence tier.}
  \label{fig:pipeline}
\end{figure}

Each case is released as a case-level record (Figure~\ref{fig:pipeline}) rather
than as a segmentation file alone, so generated labels travel with the
fields needed to audit, filter, weight, or re-threshold cases after release.
Raw-image access follows the OpenNeuro licensing and provenance accounting
described in Appendix~\ref{sec:appendix_datasheet}.

\subsection{Completeness Policy}
\label{sec:dataset:cohort}
A case is included in the hosted release only when all release evidence is
present; cases with incomplete QC-7 evidence are excluded rather than imputed,
because complete risk evidence is required for filtering, weighting, auditing,
and tiering after release. Available age, sex, scanner, field-strength, and
health-status metadata are reported as released, not imputed, and are used for
release audit rather than fairness claims;
Appendix~\ref{sec:appendix_datasheet} reports metadata coverage and provenance
accounting.

\subsection{Generated Labels and Committee Evidence}
\label{sec:dataset:labels}
OpenBrain releases two kinds of generated-label evidence for each case: a
primary generated parcellation for downstream use, and two additional generated
parcellations for risk assessment. The primary parcellation comes from
$M_1$. The committee parcellations come from $M_2$ and $M_3$, which are
architecturally distinct from $M_1$ and from each other. They are released as
reliability evidence: they are not ground truth, not manual annotations, and
not fused consensus labels. Their role is to expose agreement or disagreement
among different label generators.

OpenBrain does not claim novelty as a parcellation architecture; the release
should be read as generated whole-brain parcellations with explicit risk
fields, not per-case manual annotations. Section~\ref{sec:discussion}
discusses the upstream FreeSurfer-derived training labels and evaluation
implications.

\subsection{QC-7 risk manifest and interpretation contract}
\label{sec:dataset:confidence}
\label{sec:method:committee}
\label{sec:method:geom}
OpenBrain QC-7 is a reference-free risk manifest for generated whole-brain
parcellations. It records seven checks in three failure families. \emph{Label
geometry} checks ask whether the generated foreground has implausible spatial
structure: $Q_1$ brain-mask leakage, $Q_2$ source-relative ROI volume outliers,
and $Q_3$ distant disconnected fragments. \emph{Image-label consistency} uses
$Q_4$ within-label intensity heterogeneity to ask whether predicted anatomical
regions are internally consistent in the T1w image. \emph{Committee
disagreement} compares the three generated parcellations through $Q_5$ pairwise
committee Dice, $Q_6$ voxel-vote ambiguity, and $Q_7$ committee volume
disagreement. Appendix~\ref{sec:appendix_qc_details} gives the formal
definitions.

OpenBrain converts each raw axis value into a percentile risk. Let
$\widehat F_k$ be the empirical CDF of raw axis $Q_k$ over the QC pool, with
mid-ranks for ties. Equation~\ref{eq:rk} applies to
lower-is-better axes:
\begin{equation}
\label{eq:rk}
    r_k(x) = \widehat F_{k}\bigl(Q_k(x)\bigr), \qquad
    k\in\{1,2,3,4,6,7\}.
\end{equation}
For committee Dice, larger raw values indicate stronger agreement, so
Eq.~\ref{eq:r5} inverts the percentile:
\begin{equation}
\label{eq:r5}
    r_5(x) = 1 - \widehat F_{5}\bigl(Q_5(x)\bigr).
\end{equation}
After this conversion, all seven $r_k$ fields have the same direction: larger
values mean higher generated-label risk. These fields are ranks, not
calibrated probabilities of anatomical error.
Equation~\ref{eq:combined_risk} defines the combined case risk as an
unweighted mean:
\begin{equation}
\label{eq:combined_risk}
    R(x) = \frac{1}{7}\sum_{k=1}^{7} r_k(x).
\end{equation}

OpenBrain assigns tiers by sorting the 35{,}838 hosted release cases by $R$.
The lowest-risk 8{,}960 cases form \textsc{HighConf}, the middle 17{,}918 form
\textsc{Standard}, and the highest-risk 8{,}960 form \textsc{Flagged}. The
release stores the raw $Q_k$ values, percentile risks $r_k$, combined risk
$R$, and tier label so users can filter, weight, audit, or re-threshold cases
without rerunning label generation or committee inference.
\label{sec:method:tier}

\label{sec:dataset:uses}
The manifest also sets the scope of the release. QC-7 risks and tiers are
evidence about generated-label reliability, not manual approval labels,
calibrated correctness probabilities, demographic fairness claims, or clinical
quality guarantees. Section~\ref{sec:results} tests two reuse claims: whether
QC-7 risk ranks generated-label quality on external cohorts, and whether the
released risk fields help select training cases under a fixed budget. Main-text
evaluation uses a unified 96-label space, excluding 14 sparse or ill-defined
classes from the released 110-class parcellations; Appendix~\ref{sec:appendix_110label}
reports the corresponding sensitivity analysis.
\label{sec:dataset:96label}

\section{Validation Design}
\label{sec:method}
\label{sec:validation_protocol}

The validation protocol fixes the QC-7 axes, tier thresholds, label grouping,
and primary operating point before testing whether reference-free risk ranks
generated-label quality and helps select training cases. IXI and BrainLat
support development decisions; Mindboggle-101 remains a confirmatory
manual-label holdout.

\subsection{Frozen evaluation protocol}
\label{sec:validation:cohorts}
IXI and BrainLat choose the OpenBrain QC-7 composition, unified 96-label
evaluation space, and primary 25\% \textsc{HighConf} operating point. After
Mindboggle-101 is evaluated, no QC axis, tier threshold, label grouping, or
operating-point decision is changed. Appendix~\ref{sec:appendix_preregistration}
records these frozen choices.

\subsection{Risk ranking and training selection}
\label{sec:validation:stage_a}
\label{sec:validation:training_selection}
\label{sec:validation:stage_c}
The risk-ranking evaluation applies frozen aggregate risk $R$ to each external
cohort using only reference-free inputs. For
$K\in\{10,25,50,75,90\}$, we retain the lowest-risk $K\%$ of cases and compare
mean 96-label macro Dice against the full-cohort baseline. This tests $R$ as a
case-quality ordering; bootstrap details are in
Appendix~\ref{sec:appendix_preregistration}.

For downstream training, we draw three 1{,}000-case selections from the QC
pool: lower-risk, random, and higher-risk. These names
describe selection direction under a fixed budget, not pure tier
membership~\citep{esteban2017mriqc}. Appendix~\ref{sec:appendix_preregistration}
reports the selection rules and the manifest-tier composition.

Finally, the hosted-release audit applies the same frozen QC-7 policy to all
35{,}838 hosted release cases from 607 OpenNeuro source datasets. This audit
reports tier structure, source coverage, and available demographic/scanner
metadata as release fields. Missing fields are not imputed; the Datasheet
reports coverage and provenance accounting.

\section{Results}
\label{sec:results}

The results test two reuse questions: whether lower QC-7 risks correspond to
better generated labels on external cohorts, and whether lower-risk release
cases make better training data under the same 1{,}000-case budget. Model
provenance and the 78-label ModernSurfer reproduction anchor are in
Appendices~\ref{sec:appendix_m1} and~\ref{sec:appendix_78label}; main-text
claims use the OpenBrain 96-label evaluation space.

\subsection{Risk-ranking evaluation: the QC-7 panel ranks generated-label quality}
\label{sec:stage_a}
\begin{figure}[t]
  \centering
  \includegraphics[width=\linewidth]{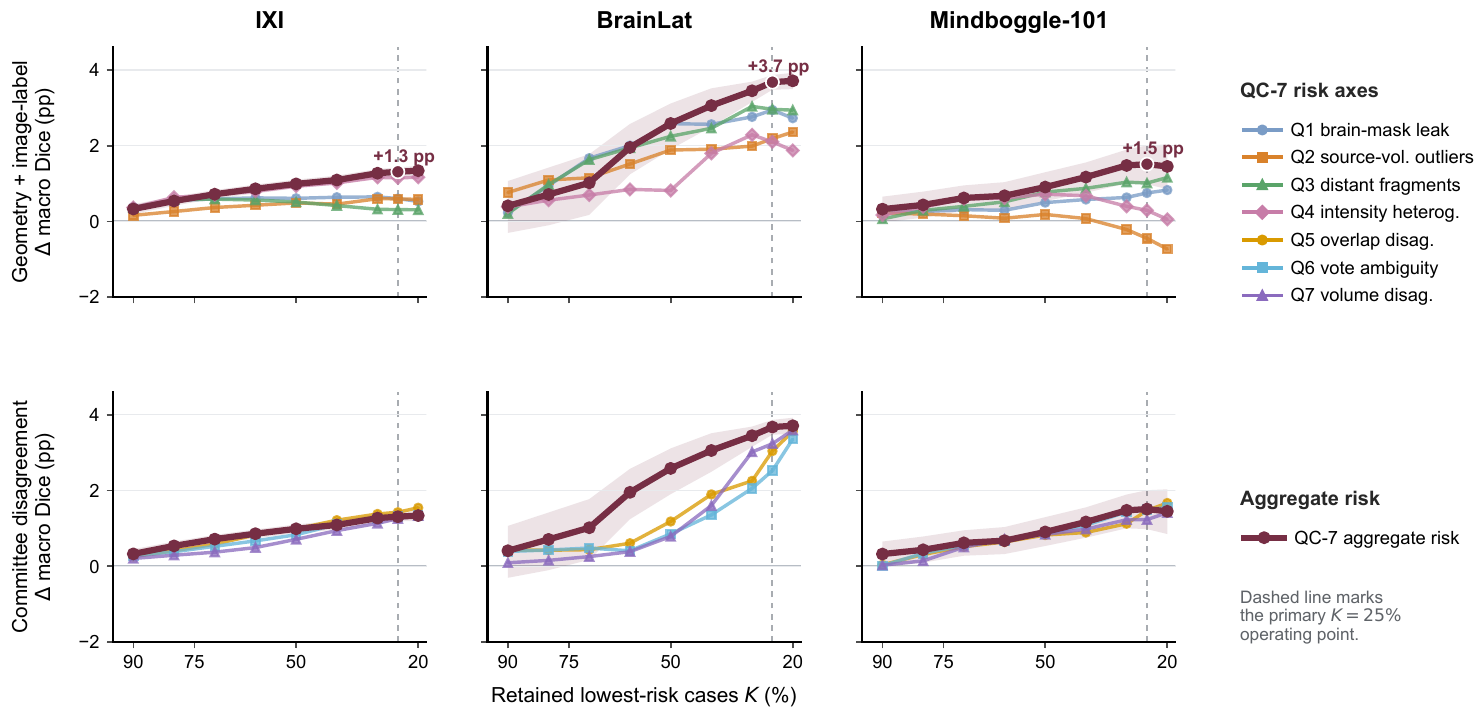}
\caption{Retention-quality curves for QC-7 risk signals. Columns are external
cohorts and rows group QC families. Curves report retained-case $\Delta$ macro
Dice after selecting the lowest-risk $K\%$ of cases.}
  \label{fig:per_qc_retention}
\end{figure}

At the primary $K=25\%$ operating point, selecting cases with the lowest
QC-7 risk raises retained-case macro Dice on all three external cohorts:
$+1.3$ percentage points on IXI ($n{=}563$), $+3.7$ pp on BrainLat
($n{=}394$), and $+1.5$ pp\ci{[+1.0, +2.0]} on the Mindboggle-101 holdout
($n{=}101$, 25 retained; Table~\ref{tab:stage_a_defA_K25}). The IXI and BrainLat
references follow the same FreeSurfer convention as $M_1$'s training labels;
Mindboggle-101 is the only manually labeled cohort and the smallest. The same
ordering advantage
persists across the reported retention grid
(Fig.~\ref{fig:per_qc_retention}). This is the core risk-ranking result:
QC-7 works as a release-time ordering signal.

\paragraph{Axes capture different failure families.}
The individual axes are not duplicates of one hidden score.
Figure~\ref{fig:per_qc_retention} shows geometry, image-label, and committee
checks peaking in different cohorts, while the aggregate is the strongest
cross-cohort summary at $K=25\%$. Partial correlations show strong coupling
only for $r_5$ committee overlap disagreement and $r_6$ vote ambiguity
($\rho{=}+0.76$; Fig.~\ref{fig:f4}A); Appendix
\ref{sec:appendix_qc_axis_corr} reports the full matrix. This coupling does
not drive the result. Recomputing $R$ without any one committee axis changes
the cross-cohort mean gain at $K{=}25\%$ by about $0.1$ pp; merging the
three into one term gives $2.04$ instead of $2.17$ pp, and $r_1$--$r_4$ alone
give $1.51$ pp (Table~\ref{tab:committee_axis_sensitivity}).

\paragraph{Sensitivity to small structures.}
Macro-Dice gains could come mostly from large cortical and white-matter
classes. On a 16-ROI subcortical subset matching the
FastSurfer~\citep{henschel2020fastsurfer} and
SynthSeg~\citep{billot2023synthseg} small-structure panels, QC-7 retention
still improves Dice on all three cohorts (Table~\ref{tab:stage_a_defA_K25}).
The gains are smaller than the full macro-Dice gains, especially on BrainLat,
so QC-7's largest benefit is not in the hardest limbic and basal-ganglia
targets.

% T-A: Stage A small-structure (Def-A, 16 ROIs) retention sensitivity at K=25%.
% Data: paper/data/T_A_small_struct_defA.csv (Def-A) + retention_sweep_v4_96label/retention_sweep_table.csv (Overall).
% Bootstrap 95% CIs on retained means. Δ in percentage points (pp).
\begin{table}[t]
  \centering
  \small
  \setlength{\tabcolsep}{4pt}
  \caption{QC-7 retention at $K{=}25\%$ for overall 96-label macro Dice and
  Def-A small-structure Dice. Values are percentage-point deviations from the
  full-cohort baseline; colored brackets are bootstrap 95\% CIs.}
  \label{tab:stage_a_defA_K25}
  \begin{tabular}{l cc}
    \toprule
    \textbf{Cohort} & \textbf{Overall $\Delta$ macro (pp)} & \textbf{Def-A $\Delta$ small-struct (pp)} \\
    \midrule
    IXI & $+1.3$\ci{[+1.2, +1.4]} & $+0.9$\ci{[+0.8, +1.0]} \\
    BrainLat & $+3.7$\ci{[+3.5, +3.9]} & $+1.5$\ci{[+1.3, +1.7]} \\
    Mindboggle-101 & $+1.5$\ci{[+1.0, +2.0]} & $+1.2$\ci{[+0.3, +2.0]} \\
    \bottomrule
  \end{tabular}
\end{table}

\subsection{Training-selection evaluation}
\label{sec:training_selection}
\begin{figure}[t]
  \centering
  \begin{tabular}{@{}c@{\hspace{0.035\linewidth}}c@{}}
    \textbf{A} & \textbf{B} \\
    \includegraphics[height=0.25\textheight]{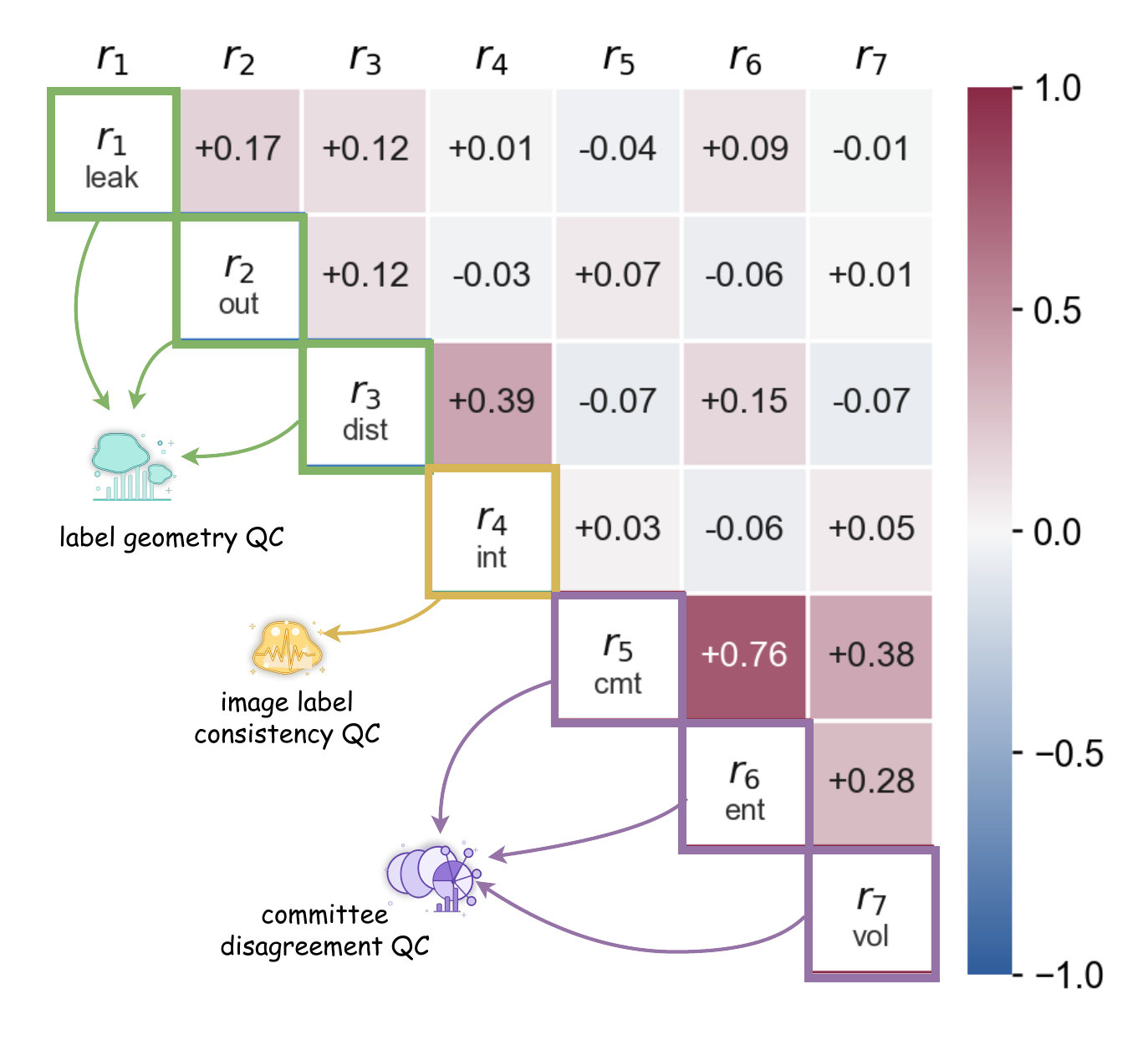} &
    \includegraphics[height=0.25\textheight,trim=0 0 145pt 0,clip]{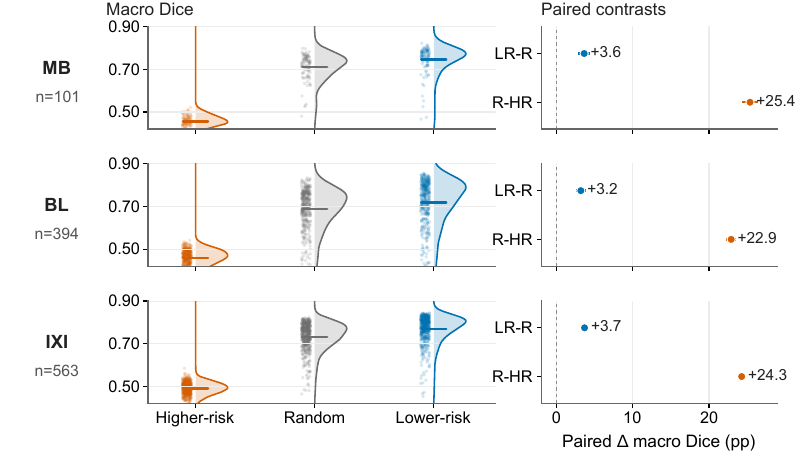}
  \end{tabular}
\caption{QC-7 axis correlation and fixed-budget training-selection results.
Panel A shows partial correlations among risk axes. Panel B compares nnU-Net
models trained on 1{,}000-case lower-risk, random, and higher-risk selections.}
  \label{fig:f4}
\end{figure}

At the same 1{,}000-case training budget, models trained on lower-risk
OpenBrain cases outperform models trained on random samples on all
three external cohorts, with paired macro-Dice gains of $+3.2$ to $+3.7$ pp
(Table~\ref{tab:stage_b_v9}; Fig.~\ref{fig:f4}B). The mirrored higher-risk
selection is an adverse-selection check. It draws heavily from the high-risk
tail, including 795 \textsc{Flagged} cases
(Appendix~\ref{sec:appendix_preregistration}, Table~\ref{tab:arm_composition}),
and performs substantially worse. These comparisons support the practical
utility of QC-guided selection under a fixed training budget. The two
selections also differ in which sources they draw from, so the gain reflects
the combined effect of risk ranking and source composition
(Appendix~\ref{sec:appendix_limitations}).

% T2: Confidence-guided lower-risk/random/higher-risk training selection -- V9 three-selection x three-cohort macro Dice (96-label).
% Data: paper/scripts/.../eval_per_case_dice.csv (mb, bl) + eval_ixi_per_case_dice.csv (ixi).
% Bootstrap 95% CIs computed over paired evaluation cases.
\begin{table}[t]
  \centering
  \small
  \setlength{\tabcolsep}{4pt}
  \caption{Fixed-budget training-selection results on three external cohorts.
  Models are trained on 1{,}000-case lower-risk, random, or higher-risk
  OpenBrain selections and evaluated with 96-label macro Dice; colored brackets
  are bootstrap 95\% CIs. The reference row gives the basis of each cohort's
  evaluation labels.}
  \label{tab:stage_b_v9}
  \begin{tabular}{l ccc}
    \toprule
    \textbf{Selection} &
    \textbf{Mindboggle-101} &
    \textbf{BrainLat} &
    \textbf{IXI} \\
    & ($n{=}101$) & ($n{=}394$) & ($n{=}563$) \\
    \textit{Reference} & manual labels & FreeSurfer-derived & FreeSurfer-derived \\
    \midrule
    Lower-risk &
    \textbf{0.746}\ci{[0.733, 0.758]} &
    \textbf{0.719}\ci{[0.709, 0.730]} &
    \textbf{0.769}\ci{[0.764, 0.775]} \\
    Random &
    0.710\ci{[0.696, 0.722]} &
    0.687\ci{[0.678, 0.697]} &
    0.733\ci{[0.727, 0.738]} \\
    Higher-risk &
    0.456\ci{[0.450, 0.461]} &
    0.458\ci{[0.454, 0.463]} &
    0.490\ci{[0.487, 0.492]} \\
    \midrule
    \multicolumn{4}{l}{\textit{Paired $\Delta$ macro Dice (pp; per-case)}} \\
    Lower-risk $-$ Random &
    $+3.6$\ci{[$+2.8$, $+4.4$]} &
    $+3.2$\ci{[$+2.6$, $+3.8$]} &
    $+3.7$\ci{[$+3.2$, $+4.2$]} \\
    Random $-$ Higher-risk &
    $+25.4$\ci{[$+24.3$, $+26.3$]} &
    $+22.9$\ci{[$+22.2$, $+23.6$]} &
    $+24.3$\ci{[$+23.8$, $+24.7$]} \\
    \bottomrule
  \end{tabular}
\end{table}

\paragraph{Where the lower-risk advantage concentrates.}
The lower-risk training advantage is smaller on the same 16-ROI subcortical
subset. Gains over random selection shrink to $+0.4$ to $+1.0$ pp
(Appendix~\ref{sec:appendix_preregistration},
Table~\ref{tab:stage_b_v9_defA}), and the BrainLat contrast is no longer
statistically distinguishable. The higher-risk deficit remains large. The
extra benefit of lower-risk selection therefore sits mostly in larger, easier
classes, while the high-risk tail remains harmful even on small structures.

\subsection{Deployment audit on the hosted release}
\label{sec:stage_c}
The same frozen policy scales to the full 35{,}838-case hosted release from
607 OpenNeuro source datasets. This descriptive audit verifies the tiered
release and treats demographic and scanner metadata as coverage fields, not
optimization targets. Provenance accounting and metadata coverage are reported
in the Datasheet and metadata audit
(Appendices~\ref{sec:appendix_datasheet}
and~\ref{sec:appendix_metadata_reliability}).

\begin{figure}[t]
  \centering
  \includegraphics[width=0.92\linewidth,trim=0 220pt 0 100pt,clip]
    {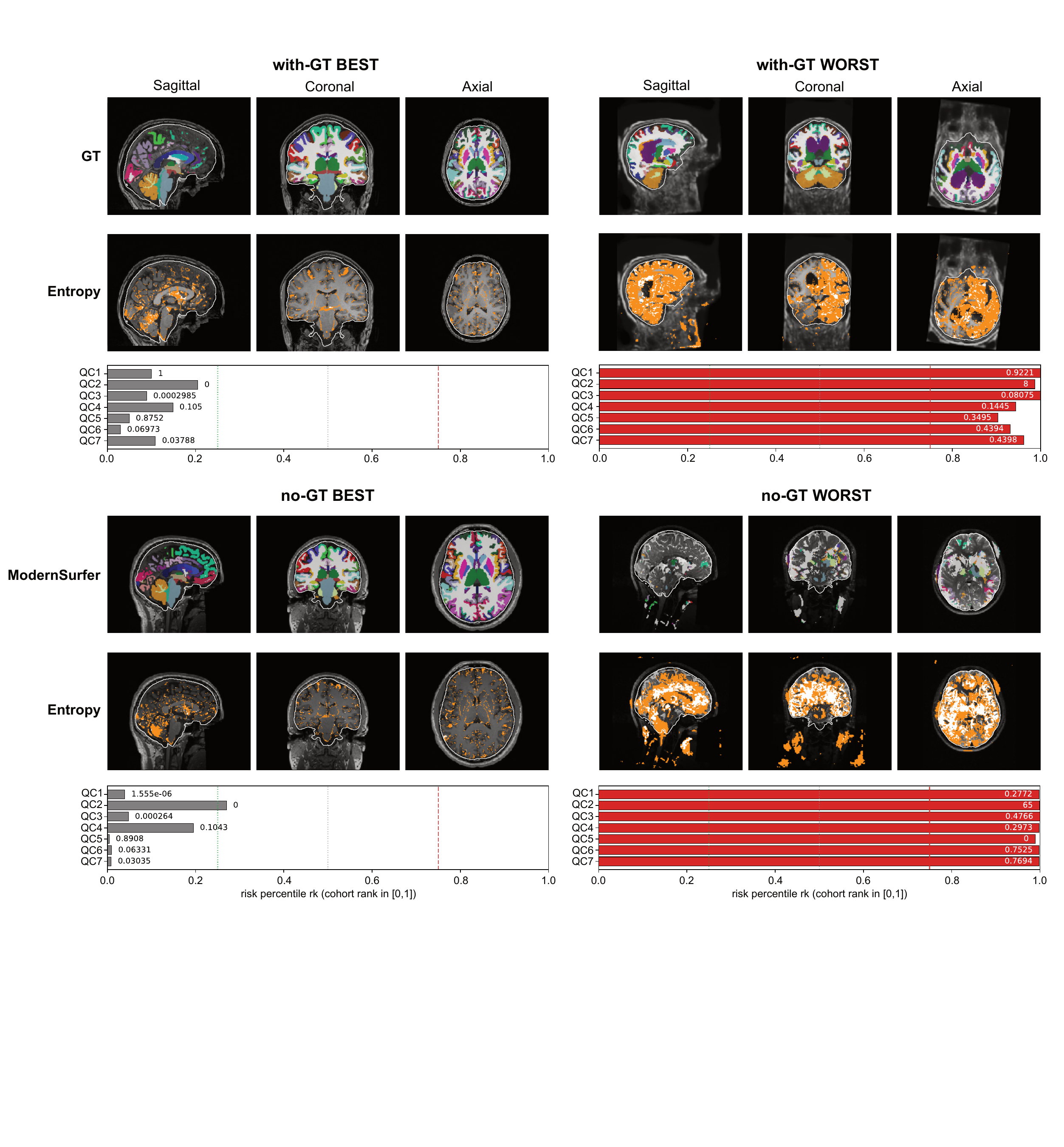}
  \caption{Qualitative audit of low- and high-risk cases. Rows separate
  manual-GT cohorts from no-GT training-pool selections; tiles show image/label
  views, entropy, QC-7 bars, and risk position.}
  \label{fig:qualitative_audit}
\end{figure}

\paragraph{Qualitative case audit.}
Figure~\ref{fig:qualitative_audit} shows what the risk ordering looks like at
case level. In the manual-GT row, the low-risk case has anatomically coherent
labels, low entropy, and low QC-7 bars, whereas the high-risk case shows visible
label disruption and committee uncertainty. The no-GT row shows the same
diagnostic pattern on release cases that users would otherwise inspect without a
manual reference. These panels are not approval labels; they show how QC-7 turns
visual failure modes into reusable evidence for trusting, auditing, or
down-weighting cases before reuse.

\section{Discussion}
\label{sec:discussion}

\paragraph{What OpenBrain contributes.}
OpenBrain's contribution is not another parcellation architecture; it is a
release format for generated labels whose reliability evidence remains visible
after download. Unlike resources that provide manual references, challenge
tasks, or pretrained outputs, OpenBrain releases a large redistributable
whole-brain generated-label corpus with per-case reliability fields
(Appendix~\ref{app:public_brain_segmentation_positioning}). QC-7's value comes
from this record: frozen checks that users can inspect before reuse.

\paragraph{What the validation shows.}
The frozen validation supports QC-7 for ranking generated-label quality on
external cohorts and guiding fixed-budget training selection. The three-group
comparison evaluates the practical utility of the overall selection policy. OpenBrain exposes both the aggregate
$R$ and the seven raw axes because the aggregate hides whether risk comes from
geometry, image-label inconsistency, or committee disagreement. The main
redundancy is within the committee family: $r_5$ captures label-overlap
disagreement, while correlated $r_6$ captures voxel-level vote uncertainty. The
formal definitions, candidate-signal history, and correlation checks are in
Appendices~\ref{sec:appendix_qc_details}, \ref{sec:appendix_b_dropped},
and~\ref{sec:appendix_qc_axis_corr}.

The appendices keep the audit trail visible through frozen selections,
110-label sensitivity, release accounting, metadata coverage, and expanded
best/worst case panels.

\paragraph{Other uses.}
OpenBrain has no disease labels. Each case record keeps its OpenNeuro source
and session identifiers, so when the source dataset publishes participant
phenotypes, users can join them under that dataset's terms, compute regional
volumes or shapes from the released parcellations, and use the QC fields to
exclude or down-weight cases before fitting a model. The phenotype labels and
their validation come from the source study.

\paragraph{What the evidence does not show.}
OpenBrain labels inherit the conventions of the upstream segmenters and their
FreeSurfer-derived training targets. Mindboggle-101 gives a manual-label
holdout, but it does not turn the release into a manual annotation dataset.
Likewise, the training-selection result shows that the manifest helps avoid a
bad high-risk tail under a fixed budget; it does not isolate every QC axis or
remove source-composition effects. Metadata fields support release audit, not
fairness claims, because coverage is incomplete and tier groups are not
metadata-matched. These boundaries are stated in the Limitations, Datasheet,
and metadata audit.

\section{Conclusion}
\label{sec:conclusion}

OpenBrain releases 35{,}838 provenance-verified generated parcellations with
per-case evidence for reliability, provenance, and reuse. The release is more
than a label collection: it defines a case-record format, exposes three
parcellation outputs, stores raw QC-7 axes and percentile risks, and assigns
confidence tiers that users can reinterpret for their own operating points.
The reusable unit is therefore the case record, not just the label file: users
can keep conservative subsets, down-weight uncertain cases, or revisit
thresholds as validation needs change.
Under frozen choices, QC-7 ranks generated-label quality on external cohorts.
The three-group comparison supports its practical value for fixed-budget
selection over random sampling and identifies a high-risk tail for closer audit. The main claim is
not that generated labels become manual annotations. It is that large
generated-label datasets should ship with visible reliability evidence.
OpenBrain gives one template for whole-brain MRI: scale, provenance, committee
evidence, QC-7 risks, validation, and explicit limits packaged together.

\begin{ack}
This work was supported in part by the National Institute on Aging under
awards U01AG079847, R01AG082721, and R01AG083039. The authors declare no
competing interests.
\end{ack}
\clearpage

\bibliographystyle{unsrtnat}
\bibliography{references}

\newpage
\appendix
\section*{Appendix Roadmap}
\label{sec:appendix_roadmap}

The appendices are organized by the evidence they support. Appendix
references below are clickable in the PDF.

\begin{center}
\small
\setlength{\tabcolsep}{6pt}
\renewcommand{\arraystretch}{1.22}
\begin{tabular}{p{0.38\linewidth}p{0.54\linewidth}}
\toprule
\textbf{Appendix} & \textbf{Evidence provided} \\
\midrule
\hyperref[sec:appendix_limitations]{\textbf{A. Limitations and scope}} &
Claim boundaries, source-composition caveats, and intended-use limits. \\
\hyperref[app:public_brain_segmentation_positioning]{\textbf{B. Benchmark positioning}} &
Positioning of OpenBrain against public brain MRI segmentation benchmarks. \\
\hyperref[sec:appendix_m1]{\textbf{C. Model provenance}} &
ModernSurfer $M_1$ training provenance and release role. \\
\hyperref[sec:appendix_b_dropped]{\textbf{D. Candidate QC signals}} &
Retained, dropped, replaced, and hard-gated candidate signals during
QC-7 development. \\
\hyperref[sec:appendix_preregistration]{\textbf{E. Holdout and training selections}} &
Fixed holdout decisions and realized composition of the fixed-budget
downstream training selections. \\
\hyperref[sec:appendix_78label]{\textbf{F. 78-label anchor}} &
FastSurfer-aligned comparison for $M_1$ and committee members. \\
\hyperref[sec:appendix_110label]{\textbf{G. Label-space sensitivity}} &
QC-7 retention check under the 110-label foreground space. \\
\hyperref[sec:appendix_datasheet]{\textbf{H. Datasheet}} &
Release composition, metadata coverage, exclusion policy, intended use,
licensing, and maintenance. \\
\hyperref[sec:appendix_qc_details]{\textbf{I. QC-7 definitions}} &
Formal definitions of raw axes, percentile risks, tiers, and
committee checks. \\
\hyperref[sec:appendix_metadata_reliability]{\textbf{J. Metadata audit}} &
Metadata-stratified QC-risk analyses and their interpretation limits. \\
\hyperref[sec:app_overview]{\textbf{K. Qualitative cases}} &
Main-text qualitative audit and detailed Mindboggle-101 worst case. \\
\bottomrule
\end{tabular}
\end{center}

\newpage

\section{Limitations and Scope}
\label{sec:appendix_limitations}

\paragraph{Generated-label circularity.}
OpenBrain labels are generated by $M_1$, which was trained against
FreeSurfer-derived labels, and two development cohorts use
FreeSurfer-derived expert-curated references. Mindboggle-101 is the
manual-label holdout that partly mitigates this circularity, but it does
not turn OpenBrain into a manual-label dataset. At $K{=}25\%$, the
Mindboggle-101 retained-case gain is $+1.5$ pp, about $40\%$ of the
BrainLat gain ($+3.7$ pp). This places the largest observed effects within
FreeSurfer-aligned evaluation conventions. Users should treat
\textsc{HighConf} cases as high-quality generated labels and should validate any
scientific conclusion that depends on fine anatomical boundaries.

\paragraph{Training-selection attribution caveats.}
The higher-risk adverse-selection deficit deserves caveats. The higher-risk
selection draws heavily from the high-risk tail, including 795 \textsc{Flagged}
cases among its 1{,}000 training cases, so its deficit relative to random
selection can absorb gross label, preprocessing, and committee-output failures;
we do not decompose that deficit by QC axis. The lower-risk and higher-risk
selections also differ in source composition as well as in risk because
percentile risk is correlated with raw image quality and source
quality varies across OpenNeuro datasets. The result shows that the released
manifest marks a tail that users should audit or avoid, not that tier labels
alone determine downstream accuracy.

\paragraph{Release completeness.}
The strict hosted release is defined by complete generated-label,
committee, metadata, QC-7, and provenance artifacts. This makes the
release directly usable for confidence-aware training and evaluation, but
it also means that hosted membership depends on preprocessing,
committee-output integrity, and provenance verification. The Datasheet
records provenance accounting and exclusion details. We also audit
QC-risk variation across available demographic and acquisition metadata;
Appendix~\ref{sec:appendix_metadata_reliability} reports the analysis and
its interpretation limits.

\paragraph{Licensing and intended use.}
OpenBrain uses OpenNeuro-hosted public images and adds derived
generated-label and QC artifacts with source provenance. The release is
intended for research on segmentation evaluation, quality-aware training,
and generated-label correction. It is not intended for clinical
decision-making, diagnosis, or patient-specific treatment planning; the
QC score is a proxy for generated-label reliability, not a clinical
quality-of-service guarantee.
\clearpage
\section{Positioning against public brain MRI segmentation benchmarks}
\label{app:public_brain_segmentation_positioning}

Table~\ref{tab:appendix_public_brain_segmentation_positioning} expands
Table~\ref{tab:related_brain_segmentation} by separating public brain MRI
segmentation resources into brain-extraction, tissue/anatomical segmentation,
and pathology-focused challenge settings. The comparison is intended to clarify
complementarity: manual anatomical and lesion benchmarks remain the appropriate
gold-standard references for their target labels, while OpenBrain targets a
different gap: large-scale T1w whole-brain derived labels with redistribution,
provenance auditing, and case-level label-quality signals.

\begin{table}[H]
\centering
\tiny
\setlength{\tabcolsep}{2.5pt}
\renewcommand{\arraystretch}{1.08}
\newcommand{\appyesmark}{\textcolor{green!60!black}{\ensuremath{\checkmark}}}
\newcommand{\appnomark}{\textcolor{red!80!black}{\ensuremath{\times}}}
\newcommand{\apppartialmark}{\textcolor{orange!85!black}{\ensuremath{\circ}}}
\caption{Detailed positioning of OpenBrain against public brain MRI
segmentation benchmarks. W: whole-brain labels; M: multi-cohort or acquisition
diversity; O: open redistribution; Q: automated label-quality scoring or
release audit.}
\label{tab:appendix_public_brain_segmentation_positioning}
\resizebox{\textwidth}{!}{%
\begin{tabular}{@{} l l l l l c c c c l @{}}
\toprule
Benchmark / resource & Scale / split & MRI / cohort & Labels & Source & W & M & O & Q & Gap vs. OpenBrain \\
\midrule
\multicolumn{10}{@{}l}{\itshape Brain extraction, tissue, and anatomical segmentation benchmarks} \\
NFBS~\citep{puccio2016nfbs} & 125 scans & NKI T1w & Brain masks & Manual correction & \appnomark & \apppartialmark & \appyesmark & \apppartialmark & Mask-only \\
CC-359~\citep{souza2018cc359} & 359 scans & Multi-vendor T1w & Brain masks & Consensus masks & \appnomark & \appyesmark & \appyesmark & \apppartialmark & Mask-only \\
Mindboggle-101~\citep{klein2012mindboggle} & 101 subj. & Multi-source T1w & DKT cortex & Manual edits & \apppartialmark & \appyesmark & \apppartialmark & \appnomark & Small; cortical-heavy \\
Hammers atlas~\citep{hammers2003atlas,gousias2008atlas} & 30 subj. & Adult T1w & 83 ROIs & Manual & \appyesmark & \appnomark & \apppartialmark & \appnomark & Small atlas cohort \\
LPBA40~\citep{shattuck2008lpba40} & 40 subj. & Adult T1w & 56 ROIs & Manual protocol & \appyesmark & \appnomark & \apppartialmark & \appnomark & Narrow healthy cohort \\
IBSR 1.5mm~\citep{frazier2007ibsr} & 18 subj. & T1w, 7--71y & 43 ROIs & Expert-guided & \appyesmark & \appnomark & \apppartialmark & \appnomark & Very small \\
MALC / NM~\citep{landman2012malc} & 30 scans; 25 subj. & OASIS T1w & WB atlas & Expert manual & \appyesmark & \appnomark & \apppartialmark & \appnomark & Small; limited redist. \\
CANDIShare~\citep{kennedy2012candishare} & 103 subj. & Pediatric T1w & CMA labels & CMA release & \appyesmark & \apppartialmark & \apppartialmark & \appnomark & Domain-specific \\
MRBrainS13~\citep{mendrik2015mrbrains13} & 20 scans; 5/15 split & Elderly 3T MRI & GM/WM/CSF & Manual reference & \apppartialmark & \apppartialmark & \apppartialmark & \appnomark & Tissue-only; small \\
MRBrainS18~\citep{kuijf2024mrbrains18} & 30 subj.; 7/23 split & T1/T1-IR/FLAIR & 8 eval. labels & Expert manual & \apppartialmark & \apppartialmark & \appyesmark & \appnomark & Compact task \\
FeTA~\citep{payette2021feta} & 50 recon.; 40/10 split & Fetal T2w & 7 tissues & Manual & \appyesmark & \apppartialmark & \apppartialmark & \appnomark & Fetal, not adult T1w \\
iSeg-2017~\citep{wang2019iseg} & 23 scans; 10/13 split & 6-mo T1/T2 & GM/WM/CSF & Manual reference & \apppartialmark & \apppartialmark & \apppartialmark & \appnomark & Infant tissue task \\
\midrule
\multicolumn{10}{@{}l}{\itshape Pathology-focused segmentation challenges} \\
BraTS 2021~\citep{baid2021brats} & 2{,}040 pts; 1251/219/570 & Glioma mpMRI & Tumor regions & Expert-approved & \appnomark & \appyesmark & \apppartialmark & \appnomark & Tumor-only task \\
ATLAS v2.0~\citep{liew2022atlas} & 1{,}271 cases; 655/300/316 & Stroke T1w & Lesion masks & Manual + QC & \appnomark & \appyesmark & \apppartialmark & \appnomark & Lesion-only task \\
ISLES 2022~\citep{hernandez2022isles} & 400 cases; 250/150 & DWI/ADC/FLAIR & Lesion masks & Expert-checked & \appnomark & \appyesmark & \apppartialmark & \appnomark & Stroke-only task \\
WMH Challenge~\citep{kuijf2019wmh} & 60 scans; 3 centers & T1/FLAIR & WMH masks & Manual reference & \appnomark & \appyesmark & \apppartialmark & \appnomark & WMH-only task \\
MSSEG 2016~\citep{commowick2018msseg} & 53 MS cases; 4 centers & Multi-modal MRI & MS lesions & 7 expert raters & \appnomark & \appyesmark & \apppartialmark & \appnomark & MS lesion task \\
ISBI 2015 MS~\citep{carass2017isbi} & 82 scans; 21/61 split & Longitudinal MRI & MS lesions & 2 raters & \appnomark & \apppartialmark & \apppartialmark & \appnomark & Small longitudinal task \\
\midrule
\textbf{OpenBrain (ours)} & \textbf{35{,}838 strict cases} & \textbf{607 OpenNeuro T1w sets} & \textbf{WB auto labels} & \textbf{Algorithmic + audit} & \appyesmark & \appyesmark & \appyesmark & \appyesmark & \textbf{Algorithmic, not manual GT} \\
\bottomrule
\end{tabular}%
}
\vspace{0.5em}
\footnotesize
OpenBrain's algorithmic labels should be interpreted as QC-aware derived labels,
not as a substitute for expert manual annotations when a gold-standard accuracy
estimate is required.
\end{table}

\clearpage
\section{Committee model \texorpdfstring{$M_1$}{M1} (ModernSurfer) training provenance}
\label{sec:appendix_m1}

The primary generated-label source $M_1$ is trained and distributed by a
separate research group. We paraphrase the training provenance below
so that OpenBrain users can trace the ModernSurfer repository and model
provenance~\citep{modernsurfer2026}; no claim of $M_1$ novelty is made
on our side.

\subsection{Training cohort}

1{,}104 T1w scans pooled from four public archives:

\begin{table}[ht]
\centering\small
\caption{ModernSurfer training cohort summarized from upstream model
provenance.}
\label{tab:m1_training_cohort}
\begin{tabular}{lcll}
\toprule
Source & $n$ & Population & Age range \\
\midrule
ABIDE I  & 100 & autism + typical & 6.5--39 y \\
CoRR     & 418 & healthy          & 6--62 y \\
ADNI     & 100 & 50 AD + 50 HC    & 55--96 y \\
SALD     & 486 & healthy          & 19--80 y \\
\midrule
total    & 1{,}104 & mixed       & 6.5--96 y \\
\bottomrule
\end{tabular}
\end{table}

Reference labels were produced by running FreeSurfer\,7.4.1 on every
scan; no manual relabeling was applied during training.
Mindboggle-101 (used as our held-out external validation set) carries
the Mindboggle protocol's manual labels plus expert repair by our
collaborators, and is not part of $M_1$'s training cohort.

\subsection{Architecture}

3D Residual-Encoder U-Net instantiated through nnU-Net v2, with six
resolution stages, feature widths (32, 64, 128, 256, 320, 320), and
residual-block counts (1, 3, 4, 6, 6, 6) from shallow to deep.
Preprocessing: foreground crop, per-volume z-score on non-zero voxels,
resample to 1$\times$1$\times$1 mm isotropic. Patch size and training
schedule follow the nnU-Net v2 defaults under this plans
configuration.

\subsection{External-validation anchor}

Appendix~\ref{sec:appendix_78label} reports the 78-label
FastSurfer-aligned external anchor used to compare our reproduction of
$M_1$ with the author-reported ModernSurfer numbers. We keep those
numbers in one place because the main-text evaluation uses the
OpenBrain-native 96-label space.

\subsection{Role in OpenBrain}

$M_1$ serves as the \emph{primary generated-label source} for all 35{,}838
hosted release cases; the generated labels we release are $M_1$'s outputs
without further editing. $M_1$ is also committee member one for the
three committee axes $Q_5, Q_6, Q_7$ (Section~\ref{sec:method:committee});
the other two members are SwinUNETR and UNETR. The committee is
intentionally architecturally diverse (CNN + Swin + ViT) so that the
disagreement signal reflects inductive-bias differences rather than
training-seed noise.

\section{Dropped candidate signals (provenance)}
\label{sec:appendix_b_dropped}

During panel development we screened additional reference-free signals on
IXI and BrainLat. Mindboggle-101 was reserved for confirmatory reporting.
The excluded signals were either weak, sign-unstable, redundant with a
retained axis, or appropriate only for pathology-specific cohorts. This
appendix records the provenance of those decisions so that users can
understand why the default release manifest is QC-7 rather than a
larger unfrozen panel.

\paragraph{Dropped-signal decisions.}
Six candidate signals were excluded from the default release panel after the
frozen development review. Brain coverage is kept only as a hard gate because it
had no stable risk-axis signal. Missing ROIs and severe fragmentation were
dropped because their signs or effect sizes were unstable once the retained
volume-outlier and distant-component axes were present. The original
small-components signal was replaced by the distant-component ratio because it
penalized true thin structures. Hemispheric asymmetry and boundary-gradient
signals were dropped because their direction depended on pathology mix or source
cohort.

Raw values for development-only signals are retained in supplementary
analysis files but are not part of the default QC-7 tiering policy.

\begin{figure}[p]
\centering
\includegraphics[width=\linewidth]{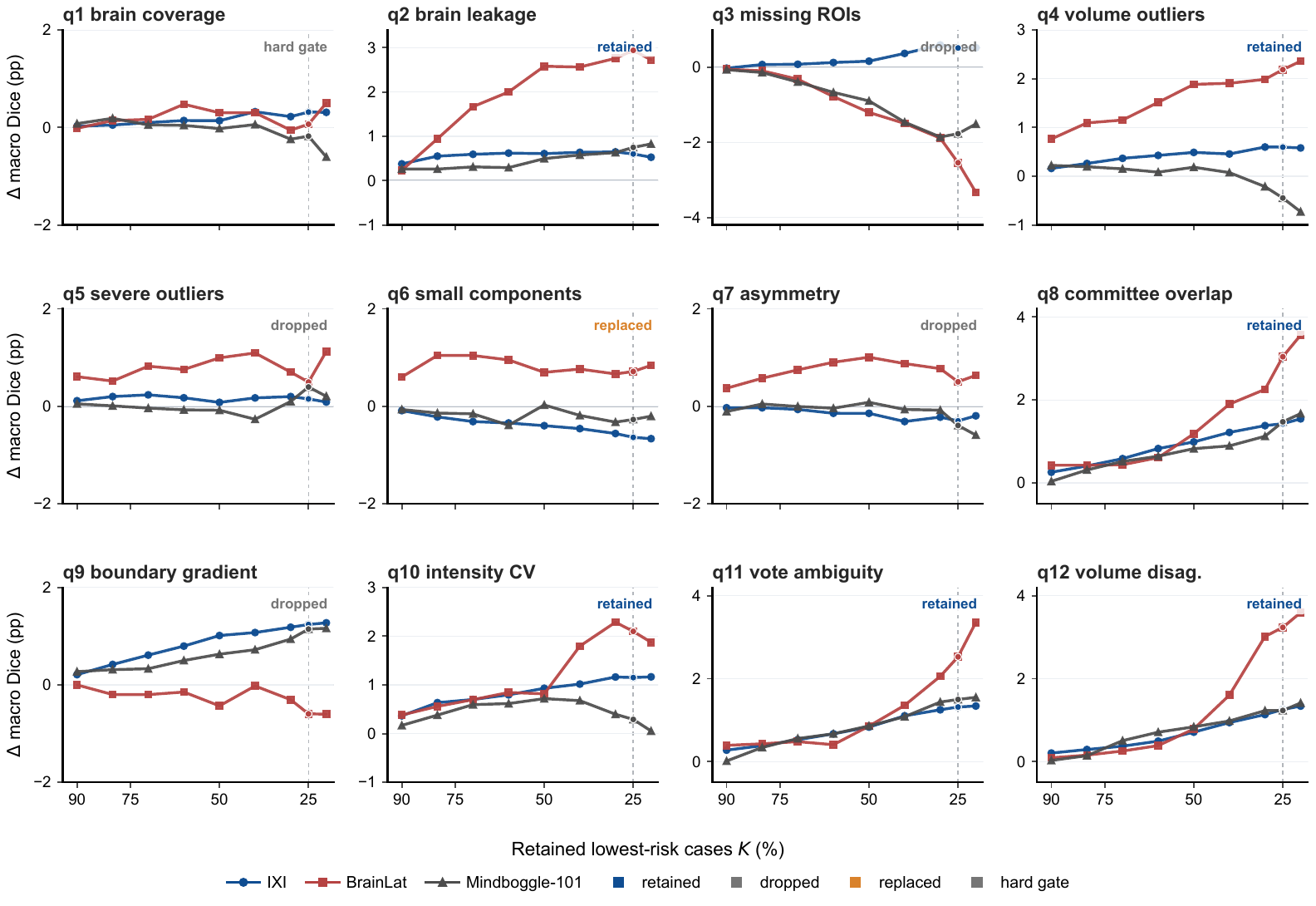}
\caption{Development provenance for the twelve original candidate QC signals.
Each panel reports retained-case $\Delta$ macro Dice after ranking by one
candidate signal. Status labels mark retained, dropped, replaced, and hard-gate
signals under the frozen QC-7 release policy.}
\label{fig:candidate_qc12_retention}
\end{figure}

Figure~\ref{fig:candidate_qc12_retention} reports the retention curves for the
twelve original candidate signals. The status labels record the frozen
development decision for each signal; they are not re-fit from the
Mindboggle-101 holdout. The retained QC-7 panel also uses the $q6p$
distant-component replacement for the original small-component signal $q6$,
which is evaluated with the released QC-7 axes in
Fig.~\ref{fig:per_qc_retention}. The candidate identifiers correspond to the
released panel as follows: $q2$ brain leakage is $Q_1$, $q4$ volume
outliers is $Q_2$, $q6p$ distant components (replacing $q6$) is $Q_3$, $q10$
intensity CV is $Q_4$, $q8$ committee overlap is $Q_5$, $q11$ vote ambiguity is
$Q_6$, and $q12$ volume disagreement is $Q_7$. $q1$ brain coverage is kept
only as a hard gate; $q3$ missing ROIs, $q5$ severe outliers, $q7$ asymmetry,
and $q9$ boundary gradient were dropped.

\section{Frozen holdout protocol and training-selection composition}
\label{sec:appendix_preregistration}

This appendix records two analysis-design details that are easy to
misread from the main text: which QC-7 decisions are fixed before the
manual-label holdout is evaluated, and how the fixed-size downstream
training selections relate to the manifest tiers.

\paragraph{Frozen holdout decisions.}
Before the Mindboggle-101 holdout is evaluated, the following are locked:
\begin{enumerate}[leftmargin=1.5em,itemsep=1pt,topsep=2pt]
    \item the 96-label unified evaluation space;
    \item the QC-7 composition and its three signal families;
    \item the primary 25\% HighConf operating point;
    \item the retention grid $K\in\{10,25,50,75,90\}\%$;
    \item the source-size gate and pooled fallback for $Q_2$.
\end{enumerate}

\paragraph{Archival mechanism.}
The frozen configuration is archived through the analysis repository tag,
the release checksum table, and the release metadata record described in the
Datasheet.

\paragraph{Holdout use.}
Mindboggle-101 is used only for confirmatory reporting of retained-case
quality and downstream model evaluation. It is not used to select axes,
drop axes, pick $K$, fit weights, choose strictness cutoffs, or validate
the $Q_2$ cohort-size gate. The 35{,}838-case hosted release manifest uses the
same frozen QC-7 panel as the holdout evaluation.

\paragraph{Training-selection composition.}
The downstream training analysis uses selection names for relative direction
under a fixed-size policy, not pure hosted-release tier membership.
Table~\ref{tab:arm_composition}
reports the realized composition of the three 1{,}000-case selections.
The selections were drawn from the 38{,}458-case QC pool. The lower-risk
and higher-risk selections rank cases by the summed ranks of $R$ and the
OpenMind image-quality score, from either end, under a per-source cap; the
random selection draws 1{,}000 cases from a random subset of 500 sources.
The release lists the 906, 935, and 931 selected cases it contains.

\paragraph{Sampling and uncertainty protocol.}
The training selections are source-aware 1{,}000-case selections from the lower-risk,
random, and higher-risk directions of the manifest, so they are not pure tier
subsets. Retention intervals use 2000 bootstrap resamples, and downstream paired
intervals use 5000 bootstrap resamples; bootstrap analyses use seed~42.

\paragraph{Training recipe.}
All three selections are trained with the same recipe: the nnU-Net v2
residual-encoder 3D full-resolution configuration, 100 epochs of 250
iterations, and the same preprocessing, optimizer, augmentation, and evaluation
code. Only the training cases differ. Each selection trains in about 6--7
hours on one NVIDIA H100 80\,GB GPU. We trained at one budget, 1{,}000
cases, and make no claim about other budgets.

\paragraph{Small-structure training sensitivity.}
The fixed-budget training result is also evaluated on the Def-A
small-structure subset. Table~\ref{tab:stage_b_v9_defA} reports these secondary
contrasts.

% T-B: Def-A small-structure (16 subcortical OB-internal IDs) Dice on V9 selections.
% Data: paper/data/T_B_small_struct_defA*.csv. Bootstrap 95% CIs.
\begin{table}[t]
  \centering
  \small
  \setlength{\tabcolsep}{4pt}
  \caption{Def-A small-structure Dice for the fixed-budget training selections
  across three external cohorts. Paired differences are reported in percentage
  points; colored brackets are bootstrap 95\% CIs.}
  \label{tab:stage_b_v9_defA}
  \begin{tabular}{l ccc}
    \toprule
    \textbf{Selection} & \textbf{Mindboggle-101} & \textbf{BrainLat} & \textbf{IXI} \\
    & ($n{=}101$) & ($n{=}394$) & ($n{=}563$) \\
    \midrule
    Lower-risk & \textbf{0.812}\ci{[0.802, 0.822]} & \textbf{0.812}\ci{[0.802, 0.822]} & \textbf{0.852}\ci{[0.846, 0.856]} \\
    Random & 0.803\ci{[0.790, 0.814]} & 0.809\ci{[0.799, 0.818]} & 0.842\ci{[0.836, 0.847]} \\
    Higher-risk & 0.557\ci{[0.544, 0.570]} & 0.537\ci{[0.529, 0.544]} & 0.582\ci{[0.576, 0.587]} \\
    \midrule
    \multicolumn{4}{l}{\textit{Paired $\Delta$ Def-A small-struct Dice (pp; per-case)}} \\
    Lower-risk $-$ Random & $+0.9$\ci{[$+0.3$, $+1.6$]} & $+0.4$\ci{[$-0.5$, $+1.2$]} & $+1.0$\ci{[$+0.5$, $+1.5$]} \\
    Random $-$ Higher-risk & $+24.6$\ci{[$+23.4$, $+25.7$]} & $+27.2$\ci{[$+26.3$, $+28.0$]} & $+26.0$\ci{[$+25.4$, $+26.6$]} \\
    \bottomrule
  \end{tabular}
\end{table}

\begin{table}[ht]
  \centering
  \small
  \setlength{\tabcolsep}{3.5pt}
  \caption{Composition of lower-risk, random, and higher-risk training selections}
  \label{tab:arm_composition}
  \resizebox{\linewidth}{!}{%
  \begin{tabular}{llll}
    \toprule
    \textbf{Selection} & \textbf{Manifest-tier composition} & \textbf{$R$ range} & \textbf{Interpretation} \\
    \midrule
    Lower-risk & 720 \textsc{HighConf} / 247 \textsc{Standard} / 33 \textsc{Flagged} & $0.082$--$0.997$ & Lower-risk selection from the low-risk end \\
    Random & 258 \textsc{HighConf} / 508 \textsc{Standard} / 234 \textsc{Flagged} & $0.123$--$0.997$ & Fixed-size random baseline without confidence ranking \\
    Higher-risk & 0 \textsc{HighConf} / 205 \textsc{Standard} / 795 \textsc{Flagged} & $0.461$--$0.997$ & Higher-risk selection from the high-risk end \\
    \bottomrule
  \end{tabular}%
  }
\end{table}

\section{Cross-method 78-label comparison (external anchor)}
\label{sec:appendix_78label}

For cross-method comparison against the author-reported
$M_1$ numbers~\citep{modernsurfer2026}, Table~\ref{tab:78label} reports
each of our three committee models on the 78-label FastSurfer-aligned
grouping used by ModernSurfer. This is an \emph{anchor table} for
cross-method orientation only; all main-text claims use the 96-label
OpenBrain-native space.
The BrainLat subset is smaller than the reported ModernSurfer subset
($n\!=\!394$ vs.\ $n\!=\!418$) because 24 cases failed OpenBrain's
preprocessing integrity checks.

\begin{table}[ht]
\centering\small
\caption{Per-committee-member mean macro Dice ($\pm$SD) on the 78-label
FastSurfer-aligned grouping.}
\label{tab:78label}
\begin{tabular}{llccc}
\toprule
Dataset & Committee model & $n$ & Mean Dice $\pm$ SD & Reported $M_1$ \\
\midrule
IXI            & $M_1$ ModernSurfer & 563 & 0.8895 $\pm$ 0.0176 & 0.8895 $\pm$ 0.0176 \\
IXI            & $M_2$ SwinUNETR    & 563 & 0.8424 $\pm$ 0.0532 & --- \\
IXI            & $M_3$ UNETR        & 563 & 0.8051 $\pm$ 0.0739 & --- \\
\midrule
BrainLat       & $M_1$ ModernSurfer & 394 & 0.8651 $\pm$ 0.0631 & 0.8798 $\pm$ 0.0282$^{\ast}$ \\
BrainLat       & $M_2$ SwinUNETR    & 394 & 0.7936 $\pm$ 0.1164 & --- \\
BrainLat       & $M_3$ UNETR        & 394 & 0.7236 $\pm$ 0.1570 & --- \\
\midrule
Mindboggle-101 & $M_1$ ModernSurfer & 101 & 0.8200 $\pm$ 0.0277 & 0.8223 $\pm$ 0.0162 \\
Mindboggle-101 & $M_2$ SwinUNETR    & 101 & 0.7940 $\pm$ 0.0503 & --- \\
Mindboggle-101 & $M_3$ UNETR        & 101 & 0.7571 $\pm$ 0.1032 & --- \\
\bottomrule
\end{tabular}

\medskip
\footnotesize
$^{\ast}$ Reported BrainLat number uses $n\!=\!418$; our pipeline
passes $n\!=\!394$ after filtering.
\end{table}

\paragraph{What the committee gap tells us.}
The $0.03$--$0.14$ macro-Dice gap between $M_1$ and the two
transformer members ($M_2, M_3$) is meaningful agreement signal: the
committee is not three copies of the same inductive bias, and the
pairwise-Dice-25\textsuperscript{th}-percentile axis $Q_5$ therefore
measures genuine architectural disagreement rather than seed noise.
A cross-architecture vs.\ cross-seed ablation would further strengthen
this claim; we defer that to future work.

\section{Label-space sensitivity (110 vs.\ 96 labels)}
\label{sec:appendix_110label}

We chose the 96-label unified space for the main-text QC retention
claims so that the committee axes $Q_5, Q_6, Q_7$ evaluate on a label
set that is consistently defined across the committee members
(Section~\ref{sec:dataset:96label}). To confirm the main-text
conclusions are not an artifact of this exclusion, we re-ran the full
retention sweep on the 110-label foreground space.

% Auto-generated by collaboration/appendix_label_space_sensitivity_table/code/build_label_space_sensitivity_table.py
\begin{table}[ht]
\centering
\small
\setlength{\tabcolsep}{4pt}
\caption{Label-space sensitivity at the primary $K\!=\!25\%$ operating point.
Effects are percentage-point changes in retained-case macro Dice relative to
the all-case baseline.}
\label{tab:label_sensitivity}
\begin{tabular}{lccccc}
\toprule
Cohort & $n$ kept & 96-label kept mean & 96-label $\Delta$ & 110-label $\Delta$ & Gap \\
\midrule
IXI & 141 & 0.9028 & +1.3 & +1.6 & +0.3 \\
BrainLat & 98 & 0.9004 & +3.7 & +3.6 & -0.0 \\
Mindboggle-101 & 25 & 0.8327 & +1.5 & +1.4 & -0.1 \\
\bottomrule
\end{tabular}
\end{table}

On all three cohorts, the sign and approximate magnitude of
QC-7's $\Delta$\,macro Dice survive the label-space change, and the
full retention curve remains positive over
$K\!\in\!\{10, 25, 50, 75, 90\}\%$. We therefore present only 96-label
numbers in the main text and report 110-label numbers here.

\section{Datasheet for Datasets}
\label{sec:appendix_datasheet}

We follow the Datasheets for Datasets template~\citep{gebru2021datasheet}.

\subsection*{Motivation}
\textbf{Purpose.} OpenBrain was created to support confidence-aware
whole-brain parcellation research at a scale beyond manual annotation and
to expose per-case quality information for generated labels.

\textbf{Creators and maintainers.} The dataset is maintained by the
OpenBrain authors through versioned releases. Each release ships a checksum
table that fixes the content of every released file.

\subsection*{Composition}
\textbf{Instances.} The submitted hosted release contains 35{,}838
provenance-verified T1w cases from 607 OpenNeuro source datasets. Each hosted
release case includes the T1w image reference, a primary generated
parcellation, committee parcellations, all seven raw QC-7 axes, percentile
risks, a combined risk score, a tier, and provenance metadata.

\textbf{Relationships.} Longitudinal repeats and multiple sessions are
represented through the BIDS source, subject, and session identifiers in
the manifest.

\textbf{Excluded candidates.} Cases are excluded from the hosted release when
source provenance, preprocessing, affine integrity checks, committee inference,
or any QC-7 axis required for the risk manifest is incomplete. Missing QC-7 or
risk fields are not imputed, because the hosted release includes only cases with
the complete provenance, parcellation, committee, QC-7, risk, tier, and license
metadata required by the release manifest.

% Auto-generated by collaboration/appendix_datasheet_provenance_table/code/build_datasheet_provenance_table.py
\begin{table}[ht]
\caption{Release provenance, exclusion accounting, and validation resources.}
\label{tab:composition_appendix}
\centering
\small
\resizebox{\linewidth}{!}{%
\begin{tabular}{lrrll}
\toprule
Resource / accounting set & Cases & Sources & Label or audit basis & Role \\
\midrule
OpenNeuro T1w records audited & 38,458 & 667 & BIDS path + source licence/provenance audit & candidate pool \\
Strict hosted OpenBrain release & 35,838 & 607 & generated WB labels + complete QC-7 manifest & released cases \\
Excluded during audit & 2,620 & 62 & manual-review, missing, non-CC0, missing/ambiguous path & not hosted \\
IXI validation & 563 & 1 & expert-curated anatomical reference & external validation \\
BrainLat validation & 394 & 1 & expert-curated anatomical reference & external validation \\
Mindboggle-101 holdout & 101 & 1 & manual labels + expert repair & confirmatory holdout \\
\bottomrule
\end{tabular}%
}
\end{table}

\textbf{Exclusion reasons.} The 2{,}620 excluded cases were removed before
QC for one of five reasons: 1{,}350 cases (43 source entries) await manual
review of their redistribution provenance; 700 cases (5 entries) came from
OpenNeuro sources that were unavailable at snapshot time; 284 cases (4
entries) had no file path despite an eligible license; 280 cases (9 entries)
carry a non-CC0 license; and 6 cases (1 entry) had an ambiguous path. We
report these counts but do not redistribute images or case identifiers for
sources without redistribution rights, and we do not claim that the release is
population-representative.

\textbf{$Q_2$ fallback.} $Q_2$ uses source-specific volume fences when a
source has at least 30 cases in the QC-fitting pool, which is counted before
release-eligibility filtering, and pooled fences otherwise. In the hosted
release, 31{,}487 cases from 311 sources use source-specific fences and
4{,}351 cases from 296 sources (12.1\%) use pooled fences. The two groups
differ in tier mix (pooled: 32.7\% \textsc{HighConf}, 44.1\%
\textsc{Standard}, 23.2\% \textsc{Flagged}; source-specific: 23.9\%, 50.8\%,
25.2\%), so the fallback is not neutral. A per-case table in the release
records which fence each case used, so users can audit, drop, or recalibrate
the pooled group.

Metadata availability and tier imbalance are summarized with the reliability
audit in Table~\ref{tab:metadata_coverage_tier_merged}; missing fields are not
imputed.

\subsection*{Collection process}
\textbf{Data acquisition.} Images are selected from public OpenNeuro source
datasets. OpenBrain adds preprocessing, SynthStrip masks, committee
predictions, generated labels, and QC manifests. Human checks guide
pipeline development, QC design, and sampled failure-case review, but
OpenBrain does not add per-case manual label edits to every release case.

\textbf{Preprocessing.} Images are resampled, skull-stripped, and
intensity-normalized before generated-label production. Raw per-axis QC
values are released alongside percentile risks and tiers.

\subsection*{Uses}
\textbf{Intended uses.} Training and evaluating brain parcellation
models, studying reference-free QC, quality-aware sampling, generated-label
correction, and active-learning style data selection.

\textbf{Unintended uses.} Clinical decision-making, diagnosis, and
patient-specific treatment planning. OpenBrain labels are generated labels with
explicit reliability information; they are not clinical-grade outputs.

\subsection*{Distribution and licensing}
\textbf{Licence.} OpenNeuro public datasets are released under CC0 terms after
publication or grace-period release. OpenBrain preserves source dataset
identifiers and citation or acknowledgement metadata in the provenance
manifest, so users can keep source attribution visible in downstream use.

\subsection*{Maintenance}
\textbf{Updates.} Updates will be issued as versioned releases. Changes
to committee checkpoints, QC axes, percentile-risk conversion, or release
membership will be reflected in the release version and its checksum table.

\section{OpenBrain QC-7 Mathematical Definitions}
\label{sec:appendix_qc_details}

This appendix gives the formal definitions behind the OpenBrain QC-7 manifest
used in Section~\ref{sec:dataset:confidence}. The panel is reference-free: it
uses the image, generated labels, brain mask, source-cohort statistics, and
committee predictions, but no target reference labels.

\subsection{Notation}
\label{sec:appendix_qc_notation}
Let $\mathcal{V}$ denote the voxel grid, $I:\mathcal{V}\to\mathbb{R}$ a T1w
image, $B:\mathcal{V}\to\{0,1\}$ a SynthStrip brain mask, and
$Y:\mathcal{V}\to\{0,1,\ldots,K\}$ the primary generated parcellation, where
label 0 denotes background. The released parcellation has $K=110$ foreground
classes. The main text evaluates QC on the unified 96-label foreground set
$\mathcal{L}\subset\{1,\ldots,K\}$ described in
Section~\ref{sec:dataset:96label}. For any label $k$, define its voxel volume
\begin{equation*}
    V_k(Y) = \left|\{v\in\mathcal{V}:Y_v=k\}\right|.
\end{equation*}
Per-ROI computations ignore labels with fewer than $v_{\min}=50$ voxels. The
three committee predictions are denoted $Y^{(1)},Y^{(2)},Y^{(3)}$, corresponding
to $M_1$, $M_2$, and $M_3$.

For a source dataset $s$, let $\mathcal{D}_s$ be the set of cases from that
source in the QC-fitting pool, which is counted before release-eligibility
filtering (Appendix~\ref{sec:appendix_datasheet}). The $Q_2$ volume fences
are fitted within $\mathcal{D}_s$ when possible; a per-case table in the
release records any fallback to pooled-cohort fences.

\subsection{Label-Geometry Axes}
\label{sec:appendix_qc_geometry}
\paragraph{$Q_1$: brain-mask leakage.}
While standard segmentation pipelines often use foreground masking as a hard post-processing step to remove structural anomalies~\citep{isensee2021nnunet}, $Q_1$ reformulates this spatial constraint into a continuous quality control metric. Brain-mask leakage measures the fraction of predicted foreground outside the brain mask:
\begin{equation*}
    Q_1(Y,B) =
    \frac{\left|\{v\in\mathcal{V}:Y_v>0\wedge B_v=0\}\right|}
         {\max\left(\left|\{v\in\mathcal{V}:Y_v>0\}\right|,1\right)}.
\end{equation*}
By normalizing against the maximum predicted foreground size, lower values indicate better label geometry without biasing against absolute brain volume.

\paragraph{$Q_2$: source-volume outliers.}
Evaluating structural volumes against normative population distributions is a well-established automated quality assurance strategy~\citep{bartz2026automated}. Following classical outlier detection validated for neuroimaging parcellation~\citep{monereo2021quality}, $Q_2$ defines out-of-distribution boundaries using Tukey's fences. For each source $s$ and label $k\in\mathcal{L}$, let
$q^{(s)}_{1,k}$ and $q^{(s)}_{3,k}$ be the first and third quartiles of
$V_k(Y)$ over cases in $\mathcal{D}_s$, and define
\begin{equation*}
    \operatorname{IQR}^{(s)}_k = q^{(s)}_{3,k} - q^{(s)}_{1,k}.
\end{equation*}
The Tukey lower and upper fences are
\begin{equation*}
    \ell^{(s)}_k = q^{(s)}_{1,k} - 1.5\operatorname{IQR}^{(s)}_k,
    \qquad
    u^{(s)}_k = q^{(s)}_{3,k} + 1.5\operatorname{IQR}^{(s)}_k.
\end{equation*}
For a case $x$ from source $s(x)$, the source-volume outlier count is
\begin{equation*}
    Q_2(Y;s(x)) =
    \left|\left\{k\in\mathcal{L}:V_k(Y)<\ell^{(s(x))}_k
    \ \vee\ V_k(Y)>u^{(s(x))}_k\right\}\right|.
\end{equation*}
Lower values are better. If $|\mathcal{D}_s|<30$, OpenBrain falls back to pooled
cohort fences for this axis and records the fallback in the manifest.

\paragraph{$Q_3$: distant fragments.}
Although evaluating the number of disconnected components is a standard topological quality check~\citep{kart2022automated,fournel2025difficulty}, raw component counts overly penalize minor one-voxel boundary artifacts. To robustly isolate geometrically implausible fragmentation, $Q_3$ introduces a distance threshold relative to a linear-radius proxy.
For each label $k\in\mathcal{L}$ with $V_k(Y)\ge v_{\min}$, let
$C_{k,1},\ldots,C_{k,J_k}$ be the 26-connected components of
$\{v:Y_v=k\}$ sorted by decreasing voxel count. Let $C_{k,1}$ be the primary
component, let $\bar c_{k,j}$ be the centroid of $C_{k,j}$, and define the
linear-radius proxy
\begin{equation*}
    \rho_k = |C_{k,1}|^{1/3}.
\end{equation*}
A non-primary component is distant if
\begin{equation*}
    \left\|\bar c_{k,j}-\bar c_{k,1}\right\|_2 > \alpha\rho_k,
    \qquad \alpha=1.5,\quad j\ge 2.
\end{equation*}
The distant-component ratio is
\begin{equation*}
    Q_3(Y) =
    \frac{\sum_{k\in\mathcal{L}}\sum_{j\ge2}
    |C_{k,j}|\,\mathbf{1}\{C_{k,j}\ \mathrm{is\ distant}\}}
    {\max\left(\sum_{k\in\mathcal{L}}V_k(Y),1\right)}.
\end{equation*}
Lower values indicate fewer geometrically implausible disconnected fragments.

\subsection{Image-Label Consistency Axis}
\label{sec:appendix_qc_image_label}
\paragraph{$Q_4$: intensity heterogeneity.}
Following standardized radiomic quality control practices~\citep{van2017computational,fournel2025difficulty}, $Q_4$ uses first-order image texture statistics to verify intra-label consistency, heavily penalizing segmentations that bleed into adjacent tissue contrasts.
For each valid label $k\in\mathcal{L}$ with $V_k(Y)\ge v_{\min}$, define
\begin{equation*}
    \mu_k = \frac{1}{V_k(Y)}\sum_{v:Y_v=k} I_v,
    \qquad
    \sigma_k = \left(\frac{1}{V_k(Y)}\sum_{v:Y_v=k}(I_v-\mu_k)^2\right)^{1/2},
\end{equation*}
and
\begin{equation*}
    \operatorname{CV}_k(I,Y) = \frac{\sigma_k}{\max(|\mu_k|,\epsilon)},
\end{equation*}
where $\epsilon>0$ prevents division by zero after intensity normalization. The
axis value is the median valid-label coefficient of variation:
\begin{equation*}
    Q_4(I,Y) = \operatorname{median}_{k\in\mathcal{L}:V_k(Y)\ge v_{\min}}
    \operatorname{CV}_k(I,Y).
\end{equation*}
Lower values indicate more homogeneous image intensities inside each predicted
ROI.

\subsection{Committee-Agreement Axes}
\label{sec:appendix_qc_committee}
\paragraph{$Q_5$: committee overlap disagreement.}
The pairwise Dice score across an ensemble is a robust, well-documented indicator of segmentation failure~\citep{zenk2025comparative}. However, standard pipelines typically aggregate these scores via the mean, which allows massive, easy-to-segment anatomical structures to dilute focal failures. To address this, $Q_5$ uses the lower quartile.
For committee members $a,b\in\{1,2,3\}$ with $a<b$, and for label
$k\in\mathcal{L}$ valid in both predictions, define
\begin{equation*}
    d_{a,b,k} =
    \frac{2\left|\{v:Y^{(a)}_v=k\}\cap\{v:Y^{(b)}_v=k\}\right|}
         {V_k(Y^{(a)})+V_k(Y^{(b)})}.
\end{equation*}
The pairwise-Dice axis is the 25th percentile of this pool:
\begin{equation*}
    Q_5(Y^{(1:3)}) =
    \operatorname{Percentile}_{25}\left(
    \{d_{a,b,k}:1\le a<b\le3,\ k\in\mathcal{L}\ \mathrm{valid}\}\right).
\end{equation*}
Higher values are better, as this lower-quartile formulation effectively isolates and emphasizes focal ROI failures.

\paragraph{$Q_6$: committee vote ambiguity.}
$Q_6$ evaluates the global spatial ambiguity of the committee's predictions using structure-wise predictive entropy, a foundational metric in Bayesian deep learning for uncertainty quantification~\citep{roy2019bayesian}.
Let
\begin{equation*}
    \mathcal{F}=\{v\in\mathcal{V}:Y^{(a)}_v>0\ \mathrm{for\ at\ least\ one}\ a\}
\end{equation*}
be the committee-union foreground. At voxel $v$, define the empirical committee
vote distribution
\begin{equation*}
    p^{(v)}_\ell = \frac{1}{3}\sum_{a=1}^{3}\mathbf{1}\{Y^{(a)}_v=\ell\},
    \qquad \ell\in\{0,1,\ldots,K\}.
\end{equation*}
The voxel entropy is
\begin{equation*}
    H_v = -\sum_{\ell:p^{(v)}_\ell>0}p^{(v)}_\ell\log p^{(v)}_\ell,
\end{equation*}
and the axis value is
\begin{equation*}
    Q_6(Y^{(1:3)}) = \frac{1}{\max(|\mathcal{F}|,1)}\sum_{v\in\mathcal{F}} H_v.
\end{equation*}
Lower values are better.

\paragraph{$Q_7$: committee volume disagreement.}
Volumetric variation across an ensemble captures size disagreements that remain hidden when masks are roughly co-located~\citep{sahlsten_application_2024,roy2019bayesian}.
For each valid label $k\in\mathcal{L}$, let
$\{V_k(Y^{(a)})\}_{a=1}^{3}$ be the three committee volumes. The structure volume coefficient of variation is defined as:
\begin{equation*}
    \operatorname{CV}^{\mathrm{vol}}_k =
    \frac{\operatorname{std}_{a=1}^{3}V_k(Y^{(a)})}
         {\max\left(\operatorname{mean}_{a=1}^{3}V_k(Y^{(a)}),1\right)}.
\end{equation*}
The volume-disagreement axis is the mean variation across valid labels:
\begin{equation*}
    Q_7(Y^{(1:3)}) =
    \operatorname{mean}_{k\in\mathcal{L}:k\ \mathrm{valid}}\operatorname{CV}^{\mathrm{vol}}_k.
\end{equation*}
Lower values are better.

\subsection{Percentile Risks and Tier Assignment}
\label{sec:appendix_qc_risk}
For each axis $k$, let $\widehat F_{k}$ be the empirical CDF of the observed
raw values $Q_k$ over the QC pool (Appendix~\ref{sec:appendix_datasheet}),
with ties assigned mid-ranks. For lower-is-better axes,
\begin{equation*}
    r_k(x) = \widehat F_{k}\bigl(Q_k(x)\bigr),
    \qquad k\in\{1,2,3,4,6,7\}.
\end{equation*}
For the higher-is-better committee-overlap axis,
\begin{equation*}
    r_5(x) = 1 - \widehat F_{5}\bigl(Q_5(x)\bigr).
\end{equation*}
Thus larger $r_k$ always denotes higher risk.

The hosted release includes only cases with all seven axis risks observed. The
combined risk is therefore
\begin{equation*}
    R(x) = \frac{1}{7}\sum_{k=1}^{7}r_k(x).
\end{equation*}

At the primary operating point, cases are sorted by increasing $R$ and assigned
tiers relative to the hosted release manifest:
\begin{equation*}
\operatorname{tier}(x)=
\begin{cases}
\textsc{HighConf}, & R(x)\ \text{is in the lowest-risk 25\%},\\
\textsc{Standard}, & R(x)\ \text{is in the middle 50\%},\\
\textsc{Flagged}, & R(x)\ \text{is in the highest-risk 25\%}.
\end{cases}
\end{equation*}
All raw $Q_k$, percentile risks $r_k$, combined risk $R$, and tier labels are
released so downstream users can re-threshold or recompose the panel without
rerunning committee inference.

\subsection{Correlation checks among QC-7 risks}
\label{sec:appendix_qc_axis_corr}
\begin{figure}[t]
  \centering
  \includegraphics[width=0.95\linewidth]{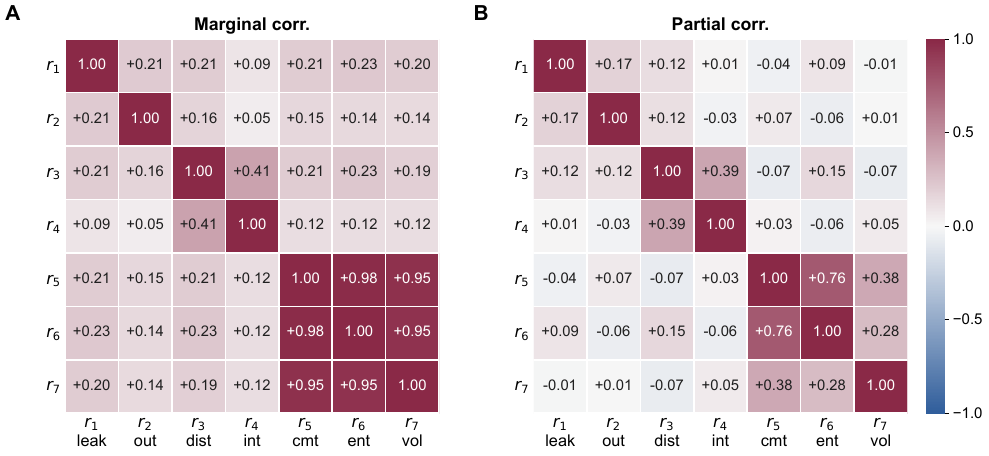}
  \caption{Marginal and partial correlations among QC-7 percentile-risk axes.
  Partial correlations condition on the other five axes. Both panels use hosted
  release cases with complete $r_1,\ldots,r_7$ values.}
  \label{fig:appendix_axes_marginal_partial_corr}
\end{figure}

Figure~\ref{fig:appendix_axes_marginal_partial_corr} gives the marginal
correlation matrix used as a check on the main partial-correlation analysis.
The committee-derived risks have high marginal correlations, but those values
shrink after conditioning on the other axes; the main exception is the
$r_5$--$r_6$ committee pair.

\paragraph{Committee-axis sensitivity.}
Table~\ref{tab:committee_axis_sensitivity} recomputes $R$ with different axis
sets on the same 1{,}058 external predictions (394 BrainLat, 563 IXI, and 101
Mindboggle-101 cases) and reports the retained-case gain at $K{=}25\%$. No
model is retrained. Removing any one committee axis changes the cross-cohort
mean by about $0.1$ pp; merging $r_5$--$r_7$ into one term gives $2.04$ pp
against $2.17$ pp for QC-7; $r_1$--$r_4$ alone give $1.51$ pp. The geometry
and image-label axes contribute on their own. We treat $r_5$--$r_7$ as one
correlated family and release every raw axis so users can set their own
weights.

\begin{table}[t]
  \centering
  \small
  \setlength{\tabcolsep}{5pt}
  \caption{Sensitivity of the $K{=}25\%$ retention gain to the committee
  axes in $R$. Values are retained-case $\Delta$ macro Dice in percentage
  points on the same external predictions; no model is retrained.}
  \label{tab:committee_axis_sensitivity}
  \begin{tabular}{l cccc}
    \toprule
    \textbf{Risk composition} & \textbf{BrainLat} & \textbf{IXI} & \textbf{Mindboggle-101} & \textbf{Mean} \\
    \midrule
    Full QC-7 ($r_1$--$r_7$) & $+3.68$ & $+1.31$ & $+1.51$ & $+2.17$ \\
    Drop $r_5$ & $+3.58$ & $+1.26$ & $+1.35$ & $+2.06$ \\
    Drop $r_6$ & $+3.61$ & $+1.33$ & $+1.51$ & $+2.15$ \\
    Drop $r_7$ & $+3.70$ & $+1.26$ & $+1.32$ & $+2.10$ \\
    Collapse $r_5$--$r_7$ to one term & $+3.61$ & $+1.15$ & $+1.34$ & $+2.04$ \\
    $r_1$--$r_4$ only & $+2.88$ & $+0.79$ & $+0.87$ & $+1.51$ \\
    \bottomrule
  \end{tabular}
\end{table}

\subsection{Committee summary statistic checks}
\label{sec:appendix_committee_aggregator}

Three of the seven checks reduce a multi-element pool to a single
per-case value: the pairwise-Dice axis pools roughly 290 (region~$\times$~pair)
values, the voxel-vote entropy axis pools every foreground voxel, and
the committee-volume axis pools one value per evaluation region. The
released panel uses the 25th percentile of the pairwise-Dice pool and
the mean of the other two pools. During panel development, we compared
these reductions with natural alternatives from
$\{\mathrm{mean},25\text{th\ percentile},\min\}$ on the external
retention sweeps. The released choices were retained because they gave
stable behavior across cohorts without making the committee axes depend
on a single extreme region or voxel. The raw axis values are released so
a downstream user who prefers a different summary statistic can
recompute the panel without rerunning inference.

\section{Metadata-Stratified Reliability Audit}
\label{sec:appendix_metadata_reliability}

We conduct a metadata-stratified audit to check whether the QC-risk
signals vary across observable subgroups. This analysis uses the random
review pool as the primary cohort because the lower-risk and higher-risk pools
were selected by QC criteria.

The audit is a reliability analysis, not a demographic fairness claim.
We treat age and sex as audit covariates, not as fairness labels. Race
and health status are reported only as metadata-coverage limitations
because their coverage is too low for subgroup error claims.

\paragraph{Baseline tier composition.}
Table~\ref{tab:metadata_coverage_tier_merged} summarizes metadata coverage and confidence-tier composition in the full 35{,}838-case release. The low-risk and flagged subsets are not metadata-matched: flagged cases have older known age on average, a larger male share, more missing age and health-status fields, a higher share of cases marked ill or pathological, and more non-standard acquisition metadata. These differences do not change the role of the tier as a QC-risk label, but they set a clear interpretation boundary. Tier comparisons should be read as reliability comparisons under the released QC policy, not as isolated demographic effects.

% Condensed from collaboration/appendix_metadata_coverage_tier_merge/data/metadata_coverage_tier_merged.csv
\begin{table}[H]
\centering
\footnotesize
\setlength{\tabcolsep}{2pt}
\renewcommand{\arraystretch}{1.12}
\caption{Metadata coverage and tier composition in the strict OpenBrain release.
Tier contrasts compare HighConf and Flagged cases as an imbalance audit, not as
demographic fairness evidence.}
\label{tab:metadata_coverage_tier_merged}
\begin{tabular}{@{}p{0.12\linewidth}p{0.15\linewidth}p{0.35\linewidth}p{0.32\linewidth}@{}}
\toprule
Field & Coverage & Full release & HighConf vs. Flagged contrast \\
\midrule
Sex & 72.9\% (26{,}114) & F/M/missing: 39.9\% / 33.0\% / 27.1\% & F: 52.0\% vs 35.0\%; M: 23.6\% vs 38.8\%; missing: 24.3\% vs 26.2\% \\
Age & 63.8\% (22{,}854) & known mean $29.2\pm15.0$ y; missing 36.2\% & known mean $26.6\pm9.2$ vs $32.0\pm18.4$ y; missing 31.7\% vs 40.4\% \\
Health & 29.0\% (10{,}409) & healthy/ill/missing: 20.3\% / 8.8\% / 71.0\% & healthy: 32.6\% vs 10.0\%; ill/path.: 6.1\% vs 10.5\%; missing: 61.3\% vs 79.4\% \\
Manufacturer & 74.9\% (26{,}833) & Siemens/Philips/GE/missing: 52.8\% / 14.2\% / 7.9\% / 25.1\% & Philips+GE: 10.1\% vs 31.1\%; missing: 31.2\% vs 16.8\% \\
Field strength & 71.0\% (25{,}446) & 3T/1.5T/7T-UHF/other/missing: 60.9\% / 4.8\% / 1.6\% / 3.7\% / 29.0\% & 1.5T: 1.3\% vs 8.9\%; 7T-UHF: 0.0\% vs 4.4\%; other: 0.0\% vs 12.7\% \\
\bottomrule
\end{tabular}
\end{table}

\begin{figure*}[t]
    \centering
    \includegraphics[width=\textwidth]{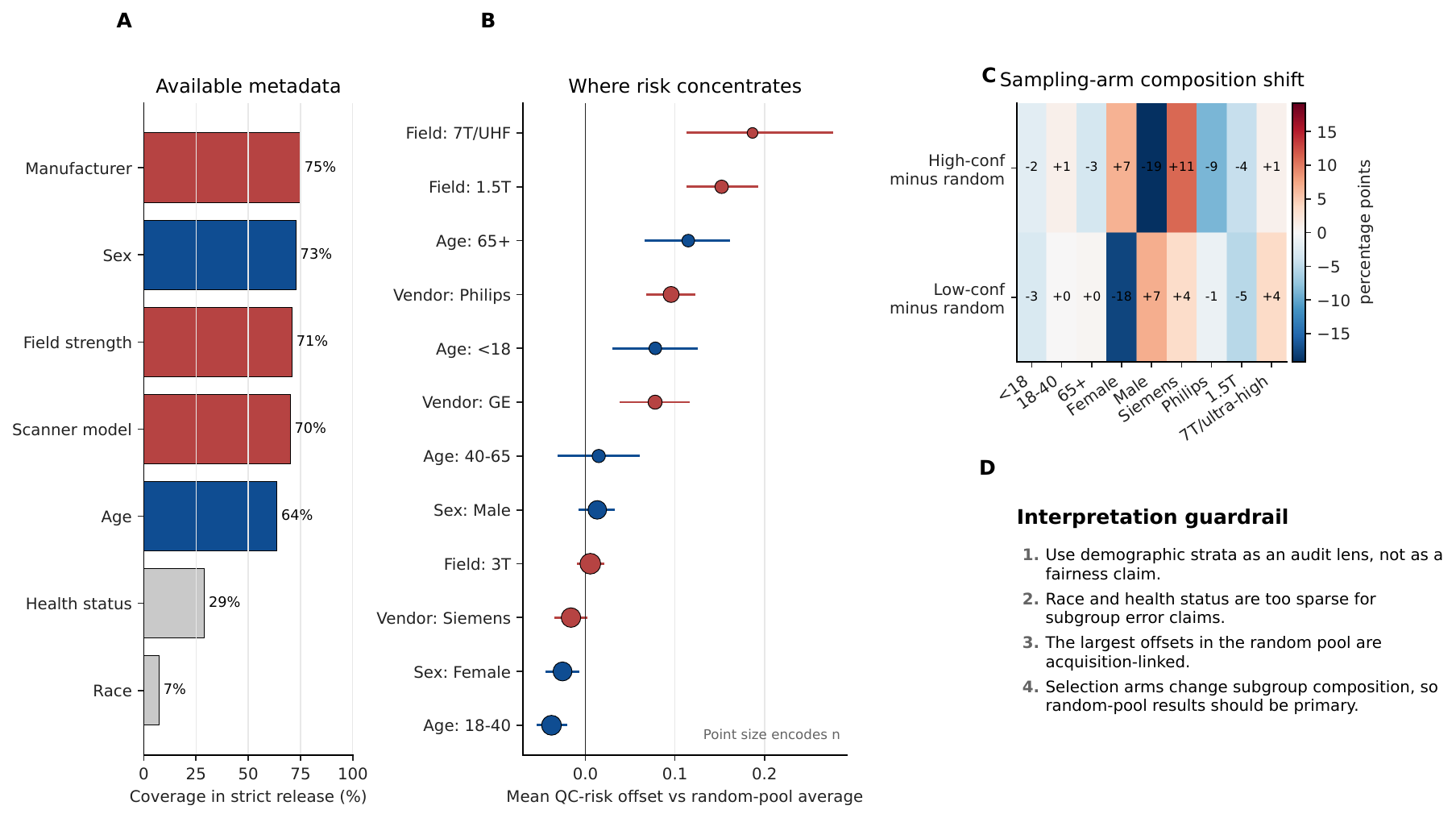}
    \caption{\textbf{Metadata-stratified QC-risk audit.}
    Panels summarize metadata coverage, subgroup mean-risk offsets, sampling-arm
    composition shifts, and the interpretation boundary for this audit.}
    \label{fig:metadata_reliability_dashboard}
\end{figure*}

\paragraph{Finding A1: coverage and mean-offset audit.}
Figure~\ref{fig:metadata_reliability_dashboard} first establishes which metadata fields are sufficiently covered, then summarizes mean QC-risk offsets and sampling-arm composition shifts. The clearest mean-risk shifts are associated with acquisition metadata: in the random review pool, 1.5T and 7T/ultra-high-field scans have higher mean QC risk than 3T scans, and Philips and GE scans also show higher risk than Siemens scans. Age groups below 18 and above 65 also have higher QC risk, but these groups are smaller and may be confounded by source dataset and acquisition protocol.

\begin{figure*}[t]
    \centering
    \includegraphics[width=\textwidth]{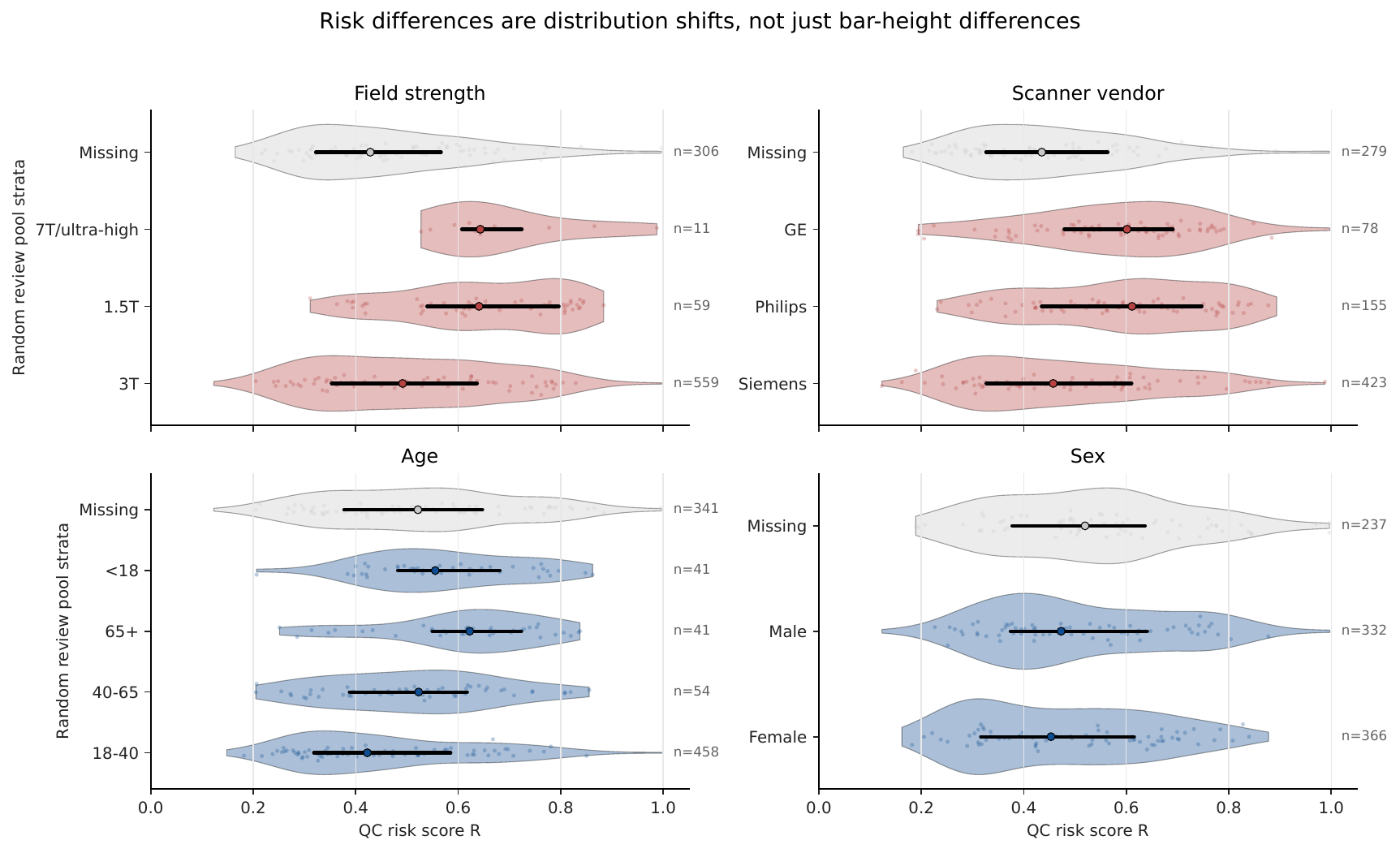}
    \caption{\textbf{Subgroup distributions of QC-risk score $R$.}
    Each row shows random-pool case-level risks; the black segment marks the
    interquartile range and the central point marks the median.}
    \label{fig:metadata_reliability_distribution}
\end{figure*}

\paragraph{Finding A2: distribution-shift audit.}
Figure~\ref{fig:metadata_reliability_distribution} tests whether the subgroup signal is only a mean-offset artifact. The raincloud views show right-shifted distributions for acquisition strata, especially field strength and scanner vendor, while retaining the full case-level spread. We therefore use the random pool for the main subgroup audit and treat lower-risk and higher-risk selection comparisons as a sensitivity check because those selections are selected by QC criteria and are not composition-matched.

Overall, this audit supports a limited conclusion. Available metadata can help identify where generated labels need closer inspection, but the current evidence does not support a demographic fairness claim. The strongest use of this analysis is to document acquisition-linked reliability differences and to motivate source-dataset-aware evaluation protocols.

\section{Qualitative case audit}
\label{sec:app_overview}

Figure~\ref{fig:qualitative_audit} provides the condensed best-vs-worst audit
in the main text. The following appendix expands the worst Mindboggle-101 case
with manual GT, committee predictions, pairwise differences, vote entropy, and
raw QC-7 values.

\clearpage
\section{Mindboggle-101 worst example}
\label{sec:app_qc_mindboggle_worst}

\begin{figure}[!htbp]
  \centering
  \includegraphics[width=\textwidth,keepaspectratio]
    {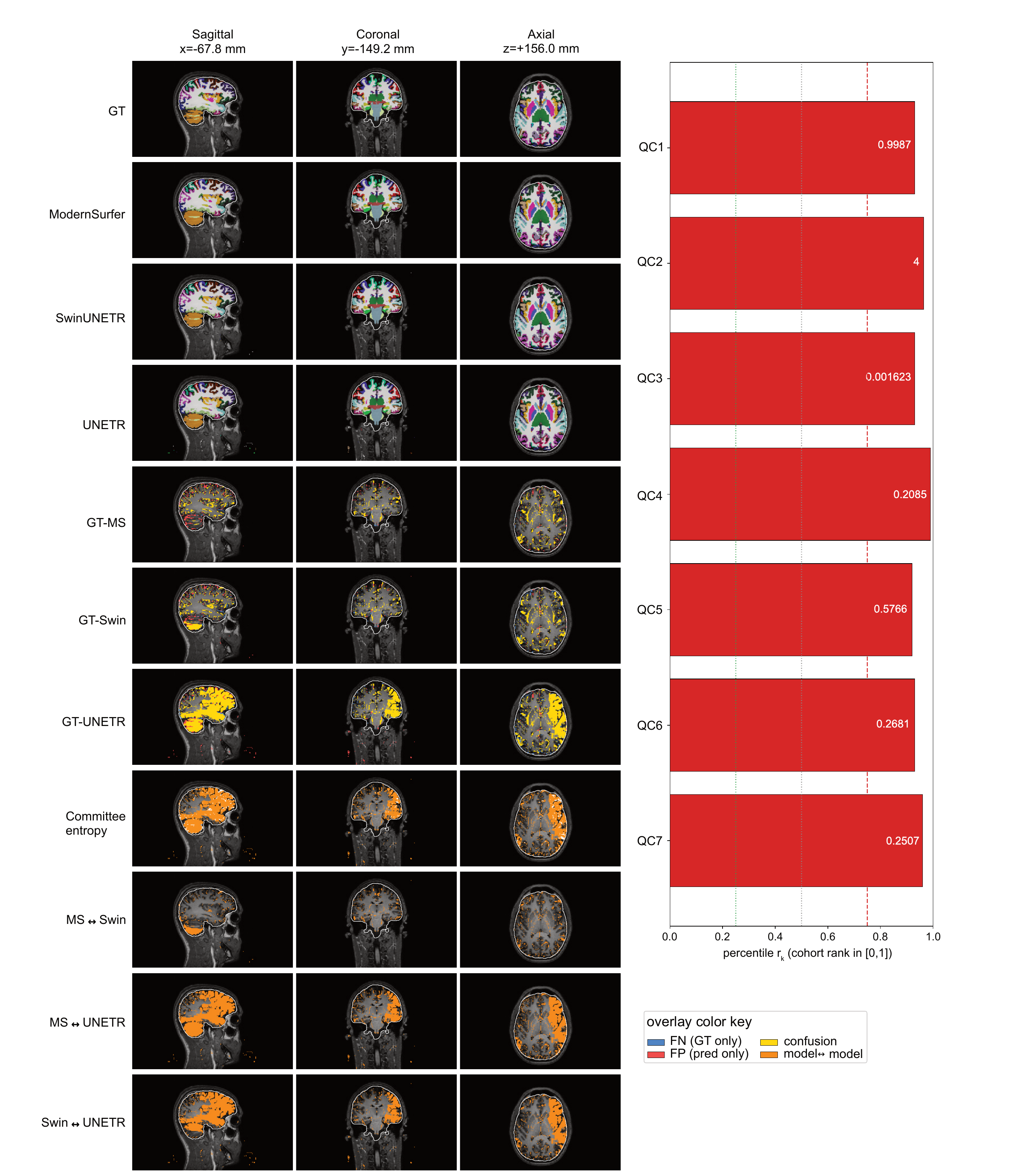}
  \caption{\textbf{Mindboggle-101 worst example.}
  Section A shows manual GT, the three committee predictions, GT--model
  differences, committee vote entropy, and pairwise model differences at the
  worst-disagreement cuts. Section B shows the corresponding QC-7 bars and raw
  $Q_k$ values.}
  \label{fig:app_mb_worst_ab}
\end{figure}
\clearpage

% NeurIPS checklist omitted from the arXiv version

\end{document}